\documentclass{article}

     \usepackage[preprint]{neurips_2025}

\usepackage[utf8]{inputenc} 
\usepackage[T1]{fontenc}    
\usepackage{hyperref}       
\usepackage{url}            
\usepackage{booktabs}       
\usepackage{amsfonts}       
\usepackage{nicefrac}       
\usepackage{microtype}      
\usepackage{xcolor}         

\usepackage{amsmath}
\usepackage{amssymb}
\usepackage{mathtools}
\usepackage{xfrac}
\usepackage{bm}
\usepackage{threeparttable}
\usepackage{tabu}
\usepackage{multirow}
\usepackage{graphicx}
\usepackage{bbding}
\usepackage{color}
\usepackage{subfigure}
\usepackage{tcolorbox}
\usepackage{colortbl}
\usepackage{geometry}

\hypersetup{
    colorlinks=true,
    citecolor=cyan,
    linkcolor=blue,
    filecolor=magenta,
    urlcolor=cyan,
    pdfpagemode=FullScreen,
    }

\newtheorem{theorem}{Theorem}[section]
\newtheorem{proposition}[theorem]{Proposition}
\newtheorem{lemma}[theorem]{Lemma}
\newtheorem{corollary}[theorem]{Corollary}

\newcounter{magicrownumbers}
\newcommand\rownumber{\stepcounter{magicrownumbers}\arabic{magicrownumbers}}

\title{Simple Convergence Proof of Adam From a Sign-like Descent Perspective}

\author{%
Hanyang Peng$^{1}$, ~~Shuang Qin$^1$ , ~~Yue Yu$^{1}$, ~~Fangqing Jiang$^1$, ~~Hui Wang$^1$, ~~Zhouchen Lin$^{2}$  \\
$^1$Peng Cheng Laboratory, Shenzhen, China\\
$^2$Peking University, Beijing, China \\
\texttt{penghy@pcl.ac.cn, qinsh@pcl.ac.cn, yuy@pcl.ac.cn} \\
\texttt{jiangfq@pcl.ac.cn, wangh06@pcl.ac.cn, zlin@pku.edu.cn} \\
}

\begin{document}

\maketitle

\begin{abstract}
Adam is widely recognized as one of the most effective optimizers for training deep neural networks (DNNs). Despite its remarkable empirical success, its theoretical convergence analysis remains unsatisfactory. Existing works predominantly interpret Adam as a preconditioned stochastic gradient descent with momentum (SGDM), formulated as $\bm{x}_{t+1} = \bm{x}_t - \frac{\gamma_t}{{\sqrt{\bm{v}_t}+\epsilon}} \circ \bm{m}_t$. This perspective necessitates strong assumptions and intricate techniques, resulting in lengthy and opaque convergence proofs that are difficult to verify and extend.
In contrast, we propose a novel interpretation by treating Adam as a sign-like optimizer, expressed as $\bm{x}_{t+1} = \bm{x}_t - \gamma_t \frac{|\bm{m}_t|}{{\sqrt{\bm{v}_t}+\epsilon}} \circ {\rm Sign}(\bm{m}_t)$. This reformulation significantly simplifies the convergence analysis. For the first time, with some mild conditions,  we prove that Adam achieves the optimal rate of ${\cal O}(\frac{1}{T^{\sfrac{1}{4}}})$ rather than the previous ${\cal O} \left(\frac{\ln T}{T^{\sfrac{1}{4}}}\right)$  under weak assumptions of the generalized $p$-affine variance  and $(L_0, L_1, q)$-smoothness, without dependence on the model dimensionality  or the numerical stability parameter $\epsilon$.
Additionally, our theoretical analysis provides new insights into the role of momentum as a key factor ensuring convergence and offers practical guidelines for tuning learning rates in Adam, further bridging the gap between theory and practice.
\end{abstract}

\section{Introduction}
 Adam  is typically formulated as follows:
\begin{equation}
\small
\begin{aligned}
&{\bm{m}}_{t} = \beta_1 {\bm{m}}_{t-1} + (1-\beta_1)\bm{g}_t, \\
&{\bm{v}}_{t} = \beta_2 {\bm{v}}_{t-1} + (1-\beta_2)\bm{g}_t^2, \\
&\bm{x}_{t+1} =  \bm{x}_t - \gamma_t \frac{\bm{m}_{t}}{\sqrt{\bm{v}_t} +\epsilon},
\end{aligned}
\label{Eq.Adam}
\end{equation}
where $\bm{g}_t= \nabla f(\bm{x}_{t};\zeta_t)$ denotes the stochastic gradient, $\gamma_t$ represents the learning rate, $\epsilon$ symbolizes the numerical stability parameter   and $\beta_1,\beta_2$ are the exponential moving average coefficients. For simplicity, bias corrections are omitted.

Currently, Adam \citep{Adam2015} has emerged as the predominant optimizer for training Transformers \citep{Transformer_2017}, particularly for state-of-the-art large language models (LLMs) \citep{GPT32020,Palm2023,Llama2023} and large vision models \citep{CLIP_2021,SAM_2023}. Notably, Adam's influence extends beyond Transformers to modern convolutional neural networks (CNNs), such as ConvNeXt \citep{ConvNext_2022,ConvNext2_2023}, where it has become the de facto choice for optimization. This is despite the traditional preference for stochastic gradient descent (SGD) \citep{AlexNet2013,ResNet2016}, which was historically considered more suitable for CNN training.

However, the theoretical convergence analysis of Adam lags behind its significant practical success. The original proof in \citep{Adam2015} was based on the convexity of the objective function but was later found to be flawed \citep{AMSGrad2018}. To address this, AMSGrad, a fixed variant of Adam, was proposed, but its theoretical analysis still relied on the convexity assumption. \cite{Adam_type_proof_2018} were the first to theoretically demonstrate that a class of Adam-type optimizers, including AMSGrad and AdaFom, converge to stationary solutions for non-convex problems. Subsequently, \cite{Simple_Adam_proof_2020} provided a simplified proof analyzing the convergence rates of vanilla Adam and Adagrad. However, their analysis required $\beta_1 < \beta_2$ and depended on the model dimensionality $d$. \cite{Prac_Adam_proof_2022} introduced practical, easy-to-check conditions to ensure the global convergence of Adam, but their proved convergence rate also heavily relied on the model dimensionality $d$. Notably, the analyses in \citep{Adam_type_proof_2018, Simple_Adam_proof_2020, Prac_Adam_proof_2022} all assumed bounded stochastic gradients. Later, \cite{Adam_proof_2022} provided a theoretical proof for random-reshuffling Adam under the weaker affine variance assumption. However, this proof achieved a slower convergence rate with an epoch-complexity bound and relies on the total number of samples. Additionally, these works assumed the conventional uniformly bounded smoothness condition, \emph{i.e.}, the $L$-smoothness condition. Recent studies, however, have shown that the $L$-smoothness assumption is inadequate for optimizing complex DNNs such as LSTMs and Transformers \citep{Clipped_SGD_Proof_2019, Generalized_signSGD_2022}. Instead, it should be relaxed to the non-uniform $(L_0, L_1)$-smoothness condition (see Section 2 for details). Recently, \cite{Adaptivity_Adam_proof_2024} analyzed random-reshuffling Adam under the $(L_0, L_1)$-smoothness assumption, but their theoretical convergence rate was still based on epoch complexity and depended on the total number of samples. \cite{Relaxed_Adam_proof_2024} demonstrated that Adam provably converges to stationary points with the optimal rate under generalized $(L_0, L_1, q)$-smoothness. However, this bound heavily relied on a large $\epsilon$, making Adam behave similarly to SGD and losing its adaptive properties. Most recently, \cite{Adam_weak_assum_2025} established the convergence rate of a simplified Adam under both the affine variance and the generalized $(L_0, L_1,q)$-smoothness assumptions. However, their results still heavily depended on the model dimensionality. A detailed comparison of these convergence analyses for Adam is provided in Table 1.


\renewcommand\arraystretch{1.1}
\begin{table*}[!tb]
  \vspace{-10pt}
\footnotesize
\centering
\caption{\small{Comparison of different convergence proofs for Adam.   ``Conv. Rate" denotes the convergence rate for approaching stationary points (\emph{i.e.}, $\Vert \nabla F(\bm{x}_T) \Vert_2 \rightarrow 0$). $T$ represents the number of iterations, $E$ the number of epochs, $d$ the model dimensionality, $n$ the total number of samples, and $\epsilon$ the numerical stability parameter.}}
\vspace{0pt}
\begin{tabu}{ l    p{2.3cm}<{\centering}p{3.0cm}<{\centering}  p{2.7cm}<{\centering}  p{2.0cm}<{\centering} }
\tabucline[1pt]{-}
References    &Noise Condition &Smooth Condition &Coeff.  Condition  &Conv. Rate\\
\vspace{-8pt} \\
\hline
 \citep{Adam_type_proof_2018}     &Bounded  Grad.  &$L$-Smooth  &$\beta_{1_t}\le\beta_1$, $\beta_{2_t} = 1-\frac{1}{t}$   &${\mathcal O}\left(\frac{{d^{\sfrac{1}{2}}}{\epsilon^{-1}}{{\rm ln}T}}{T^{\sfrac{1}{4}}}\right)$\\
  \citep{Simple_Adam_proof_2020}     &Bounded Grad. &$L$-Smooth  &$\beta_{2} < {\beta_{1}}$, $\beta_{2} = 1-\frac{1}{T}$  &${\mathcal O}\left(\frac{{d^{\sfrac{1}{2}}}{{\rm ln}({\epsilon^{-1} T})}}{T^{\sfrac{1}{4}}}\right)$\\
  \citep{Prac_Adam_proof_2022}       &Bounded Grad.  &$L$-Smooth  &$\bm{\beta_{2_t} < \sqrt{\beta_{1}}}$, $\beta_{2_t} = 1-\frac{1}{t}$  &${\mathcal O}\left(\frac{{d^{\sfrac{3}{4}}}{\rm \ln}{\epsilon^{-1}}{{\rm ln}T}}{T^{\sfrac{1}{4}}}\right)$\\
  \citep{Adam_proof_2022}      &Affine Var. &$L$-Smooth  &$\bm{\beta_{2} < \sqrt{\beta_{1}}}$, $\beta_{2} = 1-\mathcal O(\frac{1}{n^3})$  &${\mathcal O}\left(\frac{n^{\sfrac{1}{2}}{d^{\sfrac{3}{4}}}{{\rm ln}E}}{E^{\sfrac{1}{4}}}\right)$\\
  \citep{Adaptivity_Adam_proof_2024}     &Bounded Var. &$(L_0, L_1)$-Smooth  &$\bm{\beta_{2} < \sqrt{\beta_{1}}}$, $\beta_{2} = 1-\mathcal O(\frac{1}{T})$  &${\mathcal O}\left(\frac{n^{\sfrac{1}{2}}{d^{\sfrac{1}{2}}}{{\rm ln}E}}{E^{\sfrac{1}{4}}}\right)$\\
  \citep{Relaxed_Adam_proof_2024}      &sub-Gaussian Var. &\textbf{Generalized $\bm{(L_0, L_1,q)}$-Smooth}  & $\bm{\beta_{2} = 1-\mathcal O(\frac{1}{T^{\sfrac{1}{2}}})}$  &${\mathcal O}\left(\frac{\epsilon^{-2}{\rm ln}T}{T^{\sfrac{1}{4}}}\right)$ \\
  \citep{Adam_weak_assum_2025}     &Affine Var. &\textbf{Generalized $\bm{(L_0, L_1,q)}$-Smooth}  &$\beta_{2} < {\beta_{1}}$, $\beta_{2} = 1-\mathcal O(\frac{1}{T})$  &${\mathcal O}\left(\frac{d{{\rm ln}(\epsilon^{-1}T)}}{T^{\sfrac{1}{4}}}\right)$ \\
  \textbf{Corollary 3.3} \textsuperscript{1}  \tnote{*}     &\textbf{Generalized $p$-Affine Var.} &\textbf{Generalized $\bm{(L_0, L_1,q)}$-Smooth}  &{$\bm{\beta_{2} < \sqrt{\beta_{1}}}$}, $\bm{\beta_{2} = 1-\mathcal O(\frac{1}{T^{\sfrac{3}{4}}})}$  &$\bm{{\mathcal O}\left(\frac{1}{T^{\sfrac{1}{4}}}\right)}$ \\
  \vspace{-8pt} \\
\tabucline[1pt]{-}
\end{tabu}
 \begin{tablenotes}
  \footnotesize
  \item[*] 1. Compared to previous works, we establish the convergence of vanilla Adam under the weaker assumptions of generalized $p$-affine variance and  $(L_0, L_1, q)$-smoothness (see Section 2 for definitions). Furthermore, we are the first to prove that Adam achieves the optimal convergence rate of $O(\frac{1}{T^{\sfrac{1}{4}}})$ in a dimension-free and $\epsilon$-independent manner, improving upon the previous rate of ${\cal O} \left(\sfrac{\ln T}{T^{\sfrac{1}{4}}}\right)$.
  \end{tablenotes}
\label{Tab.2}
\vspace{-15pt}
\end{table*}

All existing theoretical convergence proofs for Adam are path-dependent, treating Adam as a preconditioned SGD with momentum, as initially described in \citep{Adam2015}, \emph{i.e.},$\bm{x}_{t+1} = \bm{x}_t - \frac{\gamma_t}{{\sqrt{\bm{v}_t}+\epsilon}} \circ \bm{m}_t$ where ${\sqrt{\bm{v}_t} + \epsilon}$ serves to precondition $\bm{m}_t$, introducing an effective learning rate of $\frac{\gamma_t}{{\sqrt{\bm{v}_t} + \epsilon}}$.  This preconditioned formulation not only requires strong assumptions  and intricate techniques for theoretical convergence analysis but also leads to proofs that are complex, lengthy, and difficult to verify or extend. Additionally, such theoretical analyses provide limited insights for practical optimization with Adam or for further enhancing the algorithm.

On the other hand, recent empirical evidence suggests that Adam's effectiveness may primarily stem from its sign-like property. \cite{Explian_Adam_SGD_2023} empirically demonstrates that sign descent with momentum achieves performance comparable to Adam when training Transformers, albeit without comprehensive analytical justification. Similarly, \cite{Lion2023} employs an AutoML approach to discover a highly effective optimizer, Lion, which resembles signSGD with momentum and outperforms Adam across various DNN models. More recently, \cite{Explian_Adam_SGD_2024} observed that Adam's superior performance on language models can be attributed to its sign-like property, which is particularly advantageous in addressing heavy-tailed class imbalance. However, no existing theoretical convergence proof for Adam considers its resemblance to sign descent, leaving its efficacy unexplained.

To address the aforementioned issues, we break the routine and treat Adam as a stochastic sign-like descent optimizer to analyze its convergence. Specifically, we reformulate Adam as:  $\bm{x}_{t+1} = \bm{x}_t - \gamma_t\frac{|\bm{m}_t|}{{\sqrt{\bm{v}_t}+\epsilon}} \circ {\rm Sign}(\bm{m}_t)$ where we take $\frac{|\bm{m}_t|}{{\sqrt{\bm{v}_t}+\epsilon}} $ as a single random variable.  This reformulation not only completely circumvents the aforementioned challenges but also simplifies the proof process. Moreover, the provable convergence rate of the gradient norm in expectation achieves the optimal rate under the weak assumptions of non-uniform smoothness and affine variance noise without dependency on the model dimensionality $d$ and the numerical-stability parameter $\epsilon$. Additionally, this theoretical analysis enhances our understanding of the foundations underlying Adam's success. It sheds light on why momentum improves convergence, and how to better tune hyperparameters.

 Our contributions are summarized as follow:

\begin{itemize}
  \item We pioneer the establishment of a theoretical convergence proof for vanilla Adam from the perspective of sign-like descent. This approach circumvents the intractable challenges of preconditioned settings and significantly simplifies the proof process.
  \item We are the first to prove that vanilla Adam achieves the convergence rate of { $O(\frac{1}{T^{\sfrac{1}{4}}})$}, compared to the previous { ${\cal O} \left(\frac{\ln T}{T^{\sfrac{1}{4}}}\right)$},  under the weak assumptions of generalized $p$-affine  noise and  $(L_0, L_1, q)$-smoothness along with some mild condidtions, without reliance on the model dimensionality  or the numerical stability parameter \(\epsilon\).
  \item Our theoretical convergence analysis provides the insight into the significance of momentum and provides guidance on tuning the learning rate in Adam.
\end{itemize}

\section{Preliminary }

\subsection{Notation}

Throughout this paper, we use boldface letters to denote vectors, \emph{i.e.}, \(\bm{x} \in \mathbb{R}^d\), with \(\bm{x}^{(j)}\) representing the \(j\)-th coordinate of the vector \(\bm{x}\). \(\Vert \cdot \Vert_1\) and \(\Vert \cdot \Vert_2\) denote the \(\ell_1\)-norm and \(\ell_2\)-norm, respectively. The notation \({\mathbb E} [\cdot]\) represents the expectation operator, and \(\circ\) denotes the Hadamard product (element-wise product). All divisions and square roots are computed element-wise.

In this paper, the optimizer aims to minimize the empirical risk loss of a model on a dataset, \emph{i.e.},

\begin{equation}
\min_{\bm{x} \in \mathbb{R}^d } F(\bm{x}) = {\mathbb E}_{\zeta \sim \mathcal{D}}[f(\bm{x};\zeta)] =  \frac{1}{n} \sum_{i=1}^n f(\bm{x};\omega_i),
\end{equation}
where \(\bm{x} \in \mathbb{R}^d\) and \(\zeta\) are independently and identically sampled from the dataset \(\{\omega_i: \omega_i \in \mathcal{D}, 1 \leq i \leq n \}\). For simplicity, we sometimes use \(\bm{g} = \nabla f(\bm{x};\zeta)\).

  \renewcommand\arraystretch{1.1}
 \begin{table*}[htb]
  \centering
  \small
  \begin{threeparttable}
  \begin{tabular}{p{11cm}}
  \tabucline[1pt]{-}
  \textbf{Algorithm 1.}  \textsf{ Adam}    \\
  \tabucline[0.6pt]{-}
  \small
  \rownumber: \textbf{Input}:  the momentum  $\bm{m}_0=0$ ,  $\bm{v}_0=0$, the numerical stable  constant $\epsilon$, the exponential moving average coefficient  $\beta_{1}$ and $\beta_{2}$, and the learning rate  $\gamma$. \\
  \rownumber: \textbf{for} $t=0,...,T-1$ \textbf{do} \\
  \rownumber: \hspace{8pt} Randomly sample data and compute the gradient: $\bm{g}_t \leftarrow \nabla f(\bm{x}_{t};\zeta_t)$\\
  \rownumber: \hspace{8pt} Update the momentum $\bm{m}_t$: $\bm{m}_t \leftarrow \beta_{1} \bm{m}_{t-1}+ (1-\beta_{1}) \bm{g}_t $ \\
  \rownumber: \hspace{8pt} Update the momentum  $\bm{v}$: $\bm{v}_t \leftarrow \beta_{2} \bm{v}_{t-1} + (1-\beta_{2}) \bm{g}_t^2$ \\
  \rownumber: \hspace{8pt} Compute the bias corrected $\bm{\hat{m}}_t$: $\bm{\hat{m}}_t \leftarrow  \frac{\bm{m}_{t}}{1-\beta_1^t} $ \\
  \rownumber: \hspace{8pt} Compute the bias corrected $\bm{\hat{v}}_t$: $\bm{\hat{v}}_t \leftarrow \frac{\bm{v}_{t}}{1-\beta_2^t}$ \\
  \rownumber: \hspace{8pt} Update the model parameter: $\bm{x}_{t+1} \leftarrow \bm{x}_t - \gamma\frac{\bm{\hat{m}}_t}{\sqrt{\bm{\bm{v}}_t}+\epsilon}$  \\
  \rownumber: \textbf{end for} \\
  \tabucline[1pt]{--}
  \end{tabular}
  \end{threeparttable}
  \label{Tab.1}
  \vspace{-0pt}
  \end{table*}

\subsection {Details of Adam}

To facilitate the analysis of Adam, we provide the details of Adam in Algorithm 1.

%

Following previous studies \citep{Adam_type_proof_2018, Simple_Adam_proof_2020, Adam_proof_2022, Adaptivity_Adam_proof_2024}, we omit  the bias correction in Line 6-7 of Algorithm 1 for simplicity when analyzing the convergence rate.

\subsection{Assumptions and Conditions}

To analyze the convergence rate of Adam, we list the main assumption as follows.

\textbf{Assumption A} [Bounded Infimum]. \emph{ There exists a constant $F^* > -\infty$, and the objective function follows $F(\bm{x}) \ge F^* $ for any $\bm{x} \in {\mathbb R}^d $. }

\textbf{Assumption B.1} [$L$-Smoothness] \emph{There exists a constants $L \ge 0$, and then for any $\bm{x}, \bm{y} \in {\mathbb R}^d $, the gradient of the objective function follows}
\begin{equation}
\small
 \Vert \nabla F(\bm{y}) - \nabla F(\bm{x}) \Vert_2 \le L  \Vert \bm{x} - \bm{y} \Vert_2.
\end{equation}

\textbf{Assumption B.2} [$(L_0, L_1)$-Smoothness] \emph{There exist constants $L_0, L_1\ge 0$, and then for any $\bm{x}, \bm{y} \in {\mathbb R}^d $ , the gradient of the objective function follows}
\begin{equation}
\small
 \Vert \nabla F(\bm{y}) - \nabla F(\bm{x}) \Vert_2 \le (L_0 + L_1 \Vert \nabla F(\bm{x}) \Vert_2) \Vert \bm{x} - \bm{y} \Vert_2.
\end{equation}

\textbf{Assumption B.3} [ $(L_0, L_1,q)$-Smoothness] \emph{There exist constants $L_0, L_1>0$ and $0\le q \le 1$, and then for any $\bm{x}, \bm{y} \in {\mathbb R}^d $ , the gradient of the objective function follows}
\begin{equation}
\small
 \Vert \nabla F(\bm{y}) - \nabla F(\bm{x}) \Vert_2 \le (L_0 + L_1 \Vert \nabla F(\bm{x}) \Vert_2^q) \Vert \bm{x} - \bm{y} \Vert_2.
\end{equation}

When $q = 1$, the generalized $(L_0, L_1, q)$-smoothness (Assumption B.3) is reduced to the $(L_0, L_1)$-smoothness (Assumption B.2). When $L_1 = 0$ or $q = 0$, the generalized $(L_0, L_1, q)$-smoothness (Assumption B.3) is reduced to the standard $L$-smoothness (Assumption B.1).  $(L_0, L_1)$-smoothness was originally defined in \citep{Clipped_SGD_Proof_2019} as a bound on the second-order Hessian function. Following \cite{Imprv_clipping_SGD_2020}, we reformulate the $(L_0, L_1)$-smoothness as an affine form of the gradient norm for first-order differentiable functions. Subsequently, \cite{Relaxed_Adam_proof_2024} first introduced the generalized $(L_0, L_1, q)$-smoothness to analyze the convergence of Adam, followed by \cite{Separa_Adam_proof_2024} and \cite{Adam_weak_assum_2025}.

\textbf{Assumption C.1 }[ Bounded Variance]. \emph{There exists a positive constant $\sigma_0 >0$, and then for any  $\bm{x}_t\in {\mathbb R}^d $, the noisy gradient of the objective function obeys}
\begin{equation}
\small
{\mathbb E} [\nabla f(\bm{x};\zeta)]= \nabla F(\bm{x}), ~~~~~~ {\mathbb E} [\Vert \nabla f(\bm{x};\zeta) - \nabla F(\bm{x}) \Vert_2^2] \le \sigma_0^2.
\end{equation}

\textbf{Assumption C.2 }[  Affine Variance]. \emph{There exist constants $\sigma_0, \sigma_1  \ge 0$, and then for  $\bm{x}\in {\mathbb R}^d $, the noisy gradient of the objective function obeys}
\begin{equation}
\small
{\mathbb E} [\nabla f(\bm{x};\zeta_t)]= \nabla F(\bm{x}), ~~~~~~ {\mathbb E} [\Vert \nabla f(\bm{x};\zeta_t) - \nabla F(\bm{x}) \Vert_2^2] \le \sigma_0^2 + \sigma_1^2 \Vert \nabla F(\bm{x}) \Vert_2^2.
\end{equation}

\textbf{Assumption C.3 }[  $p$-Affine Variance]. \emph{There exist constants $\sigma_0, \sigma_1 \ge 0$ and $0\le p \le 2$, and then $\bm{x}\in {\mathbb R}^d $ at any time, the noisy gradient of the objective function obeys}
\begin{equation}
\small
{\mathbb E} [\nabla f(\bm{x};\zeta_t)]= \nabla F(\bm{x}), ~~~~~~ {\mathbb E} [\Vert \nabla f(\bm{x};\zeta_t) - \nabla F(\bm{x}) \Vert_2^2] \le \sigma_0^2 + \sigma_1^2 \Vert \nabla F(\bm{x}) \Vert_2^p.
\end{equation}

When $p = 2$, the $p$-affine variance (Assumption C.3) is reduced to the affine variance (Assumption C.2). When $\sigma_1 = 0$ or $p = 0$, the  $p$-affine variance (Assumption C.3) is reduced to the bounded variance (Assumption C.1). The affine variance (Assumption C.2) was originally studied in \citep{SGD_affine_var_2000} to analyze the convergence behavior of SGD. It was later applied to analyze AdaGrad \citep{AdaGrad-Norm_affine_var_2022, AdaGrad_affine_var_2023} and a simplified Adam \citep{Separa_Adam_proof_2024}.

 To the best of our knowledge, Assumption B.3 and Assumption C.3 are the weakest assumptions for analyzing the convergence of Adam among the existing literatures.

 In addition, we also list the following required conditions.

  \textbf{Condition 1}  \emph{At any $t$-th iteration in {Algorithm 1},  the gradients satisfy $\sqrt{ \frac{1}{T}\sum_{t=0}^{T-1} \Vert \nabla F(\bm{x}_t) \Vert_2^2} \le \frac{C_0}{T}\sum_{t=0}^{T-1} \Vert \nabla F(\bm{x}_t) \Vert_2$ with $ 1 \le C_0 \ll \sqrt{T} $.}

 \textbf{Condition 2}  \emph{At any $t$-th iteration in {Algorithm 1},  the coordinates of the update in Adam , i.e., $u_t = \frac{\vert \bm{m}_t^{(j)}\vert}{\sqrt{\bm{v}_t^{(j)}}+\epsilon}(j\in[d])$ are independently and identically distributed (i.i.d). }

 \textbf{Condition 3} \emph{ At any $t$-th iteration in {Algorithm 1}, the gradient satisfies $\Vert \nabla F(\bm{x}_t) \Vert_1 =  \frac{\sqrt{d}}{{C_1}} \Vert \nabla F(\bm{x}) \Vert_2$ with $ 1 \le C_1 \ll \sqrt{d} $.}

Condition 1 is easily satisfied when the gradients $\Vert \nabla F(\bm{x}_t) \Vert_2$ decrease at a rate of ${\cal O} \left(\frac{1}{t^{\alpha}}\right)$ for all $0 < \alpha < \frac{1}{2}$, and the number of iterations is sufficiently large. We summarize this in the following proposition.

 \begin{proposition}
 If $\Vert \nabla F(\bm{x}_t) \Vert_2$ decreases at  the rate of ${\cal O } \left(\frac{1}{t^{\alpha}}\right)$ for $0\le \alpha < \frac{1}{2}$ and $T \ge 8$, then it holds that
 \begin{equation}
 \small
 \frac{\frac{1}{T}\sum_{t=0}^{T-1} \Vert \nabla F(\bm{x}_t) \Vert_2^2}{ \left(\frac{1}{T}\sum_{t=0}^{T-1} \Vert \nabla F(\bm{x}_t)\Vert_2\right)^2} \le {\cal O } \left(\frac{2(1-\alpha)^2}{1-2\alpha}\right).
 \end{equation}
 \end{proposition}

\cite{SGD_lower_bound_2023} has demonstrated that the optimal convergence rate of $\Vert \nabla F(\bm{x}_t) \Vert_2$ in non-convex stochastic optimization is $O\left(\frac{1}{T^{\sfrac{1}{4}}}\right)$, which lies in the range $[0, \frac{1}{2})$. We choose a sufficiently large $T$ in practice, meaning $T$ is greatly larger than $\frac{2(1-\alpha)^2}{1 - 2\alpha}$. Therefore, Condition 1 commonly holds in practice.

\begin{figure*}
 \vspace{-0pt}
  \small
  \centering
  \subfigure[\footnotesize ResNet-50, Layer\#32]{
		\begin{minipage}[b]{0.485\textwidth}
			\includegraphics[width=0.55\textwidth]{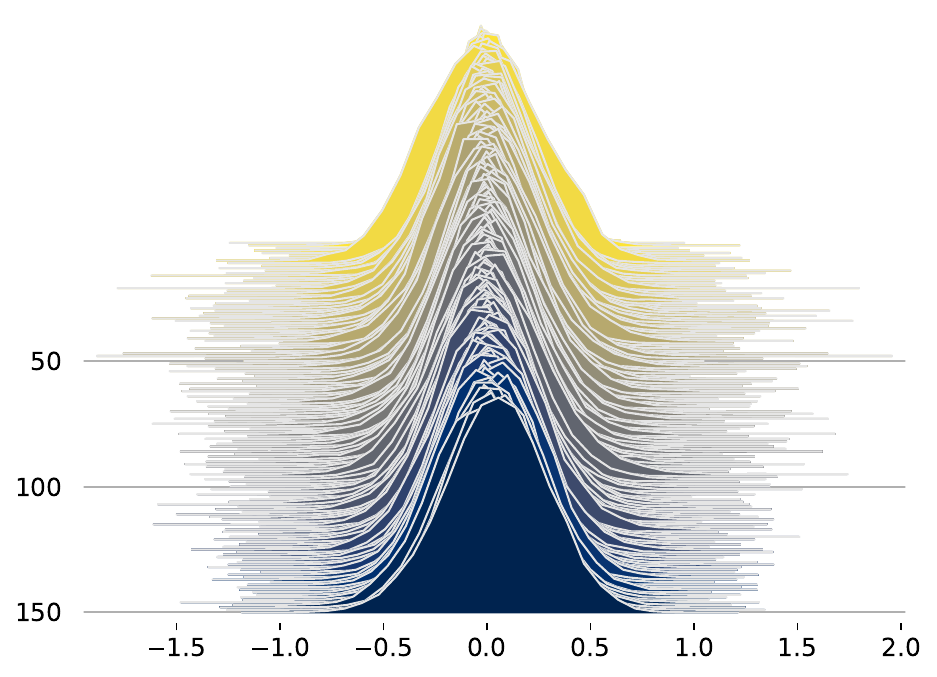}
            \includegraphics[width=0.42\textwidth]{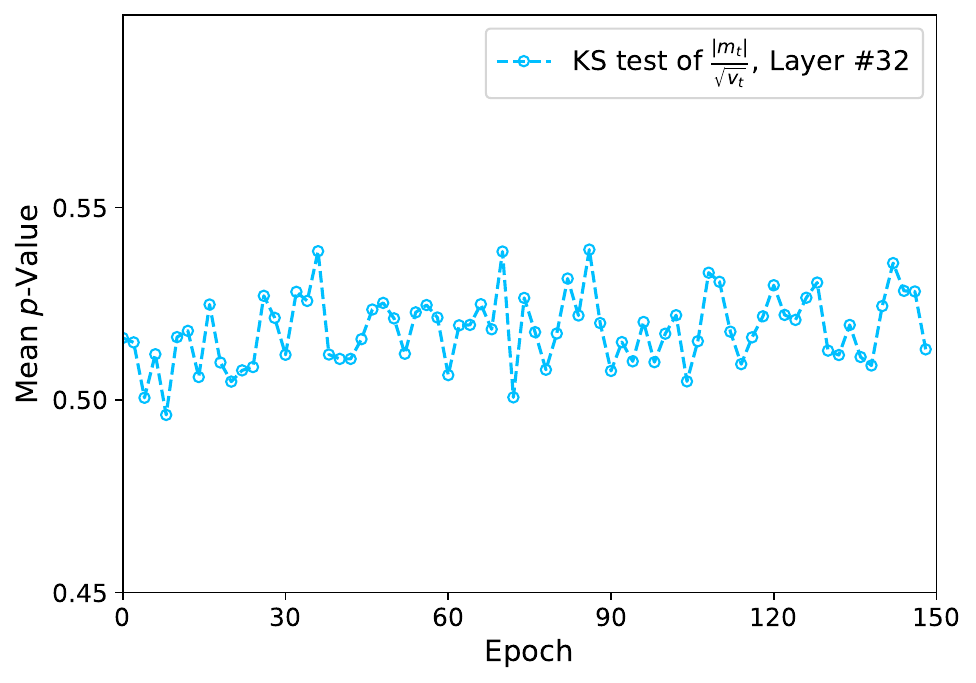}
		\end{minipage}
	}
\subfigure[\footnotesize GPT-2 (350M), Layer\#3]{
		\begin{minipage}[b]{0.485\textwidth}
			\includegraphics[width=0.55\textwidth]{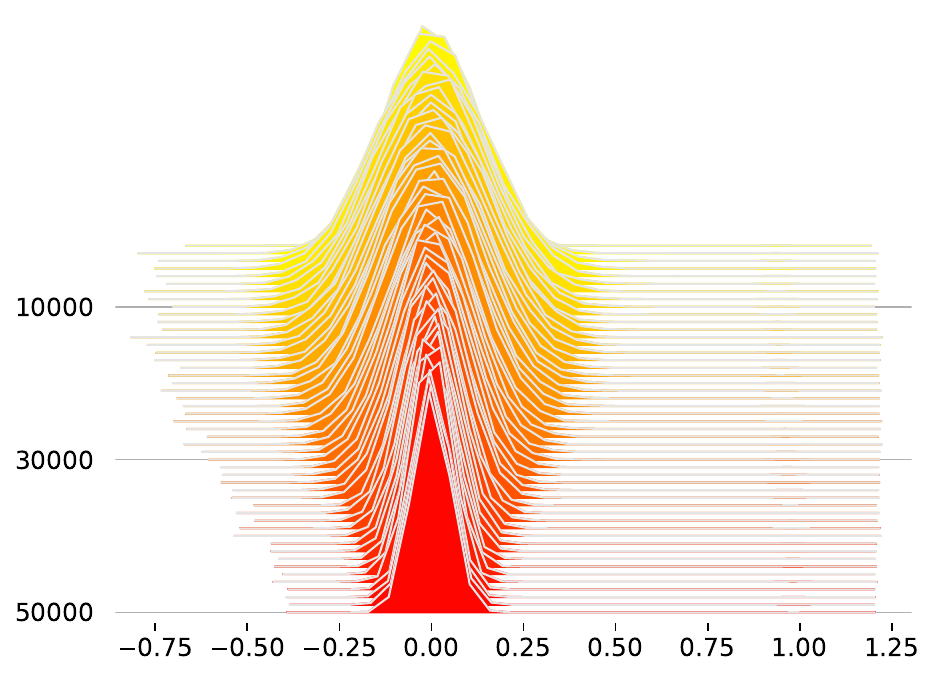}
            \includegraphics[width=0.42\textwidth]{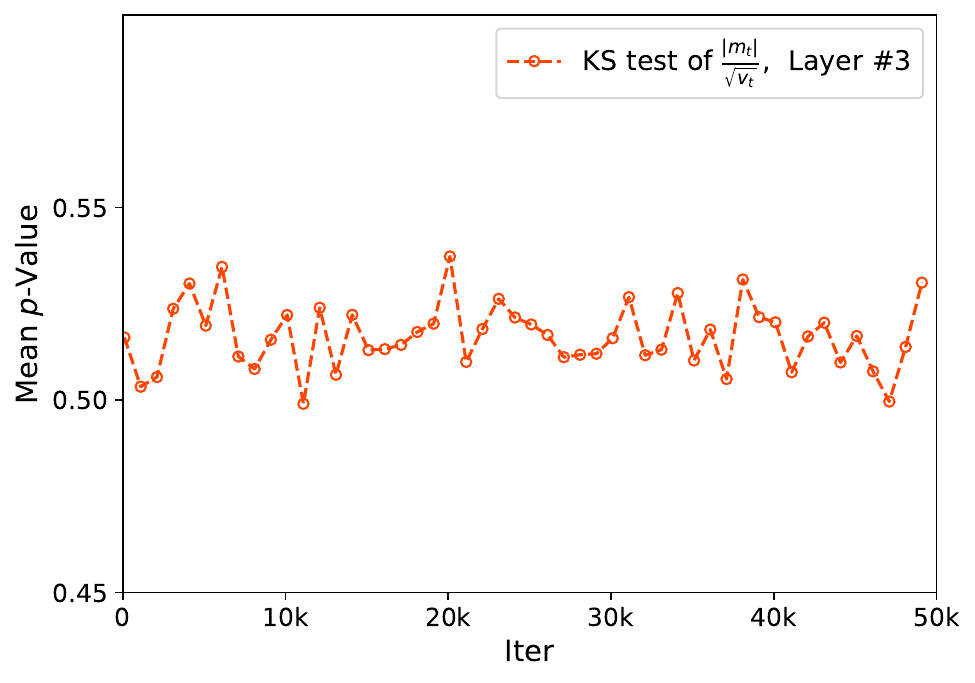}
		\end{minipage}
	}

  \caption{\small{The distribution an the two-sample Kolmogorov-Smirnov test for $\sfrac{\bm{m}_t^{(j)}}{\sqrt{\bm{v}_t^{(j)}}}$ across coordinates of (a) Layer\#32.conv.weight in ResNet-50 during training with Adam on ImageNet for 150 epochs, and (b) Layer\#3.self-attention.in-proj-weight in GPT-2 (350M) during training with Adam on OpenWebText for 5,000 iterations. In this test, two groups of 10,000 samples were uniformly drawn from all coordinates of the layer, and these groups were used to run the two-sample KS test. We repeated this procedure 1,000 times and reported the mean $p$-value. The $p$-values significantly exceed the 0.05 threshold, strongly indicating that the values of $\sfrac{\bm{m}_t^{(j)}}{\sqrt{\bm{v}_t^{(j)}}}$  are independently drawn from the identical distribution. } }
  \label{Figure.1}
\end{figure*}

 {Condition 2} commonly holds in practice, and we empirically validated it in our experiments, as shown in Figure \ref{Figure.1}. Specifically, we employed Adam to train ResNet-50 on ImageNet and GPT-2 (350M) on OpenWebText. During training, we recorded $\sfrac{\bm{m}_t^{(j)}}{\sqrt{\bm{v}_t^{(j)}}}$ for each coordinate in certain layers. To verify whether $\sfrac{\vert \bm{m}_t^{(j)}\vert}{\sqrt{\bm{v}_t^{(j)}}}$ for each coordinate is drawn from the same distribution, we used the two-sample Kolmogorov-Smirnov (KS) test.
In this test, two groups of 10,000 samples were uniformly drawn from all coordinates of the layer, and these groups were used to run the two-sample KS test. We repeated this procedure 1,000 times and reported the mean $p$-value. As illustrated in Figure \ref{Figure.1}, the mean $p$-value is significantly larger than the significance level of 0.05, strongly suggesting that $\sfrac{\bm{m}_t^{(j)}}{\sqrt{\bm{v}_t^{(j)}}}$ for each coordinate in the layers is independently drawn from an identical distribution.

\begin{figure*}
 \vspace{-0pt}
  \small
  \centering
\subfigure[\footnotesize ResNet-50, Layer\#32]{
		\begin{minipage}[b]{0.48\textwidth}
			\includegraphics[width=0.55\textwidth]{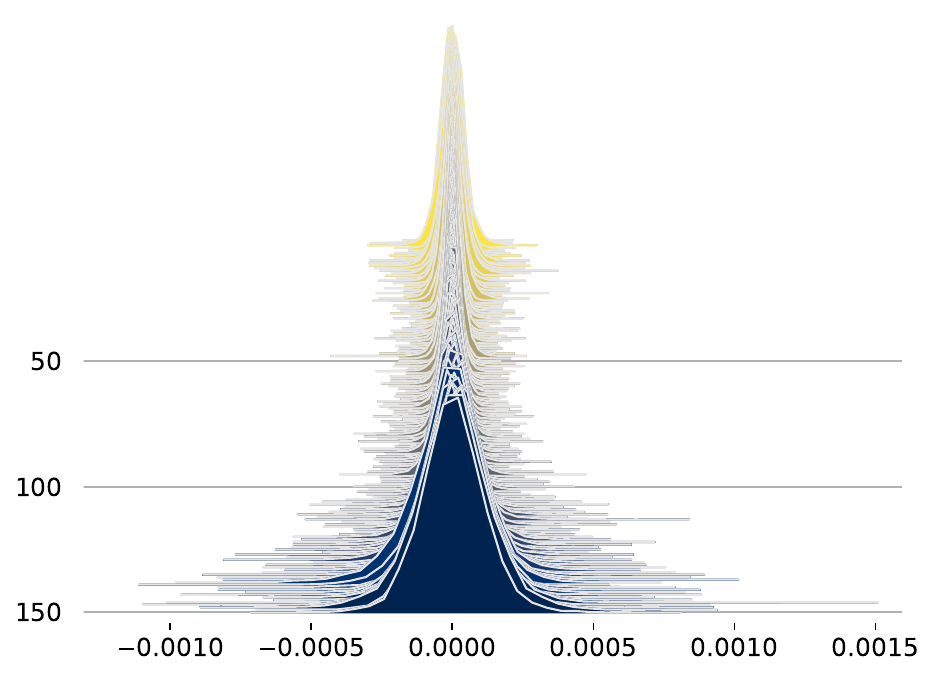}
            \includegraphics[width=0.42\textwidth]{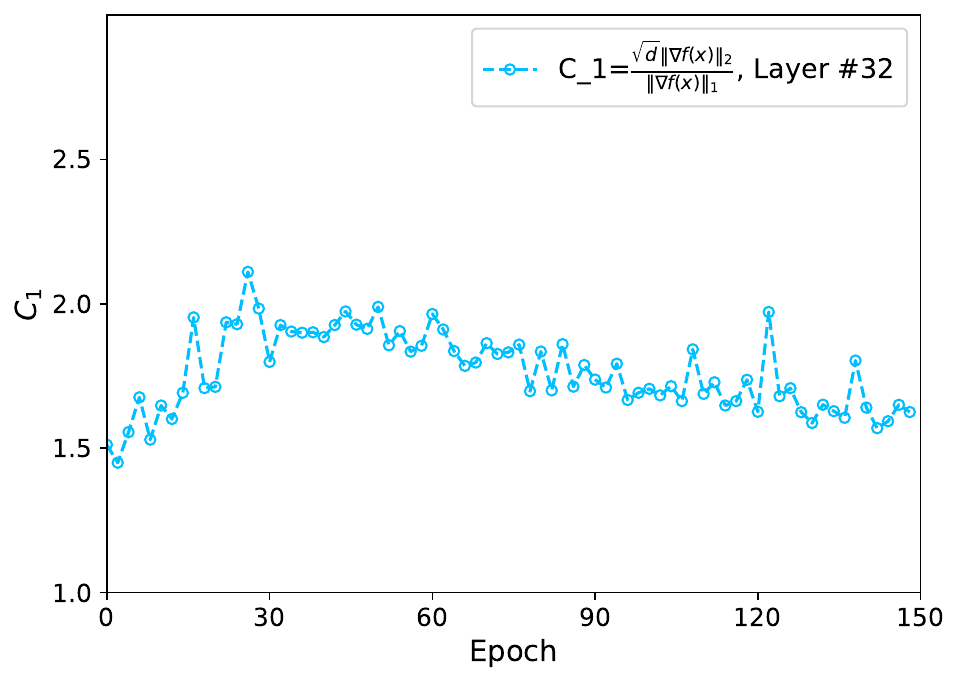}
		\end{minipage}
	}
\subfigure[\footnotesize GPT-2 (350M), Layer\#3]{
		\begin{minipage}[b]{0.48\textwidth}
			\includegraphics[width=0.55\textwidth]{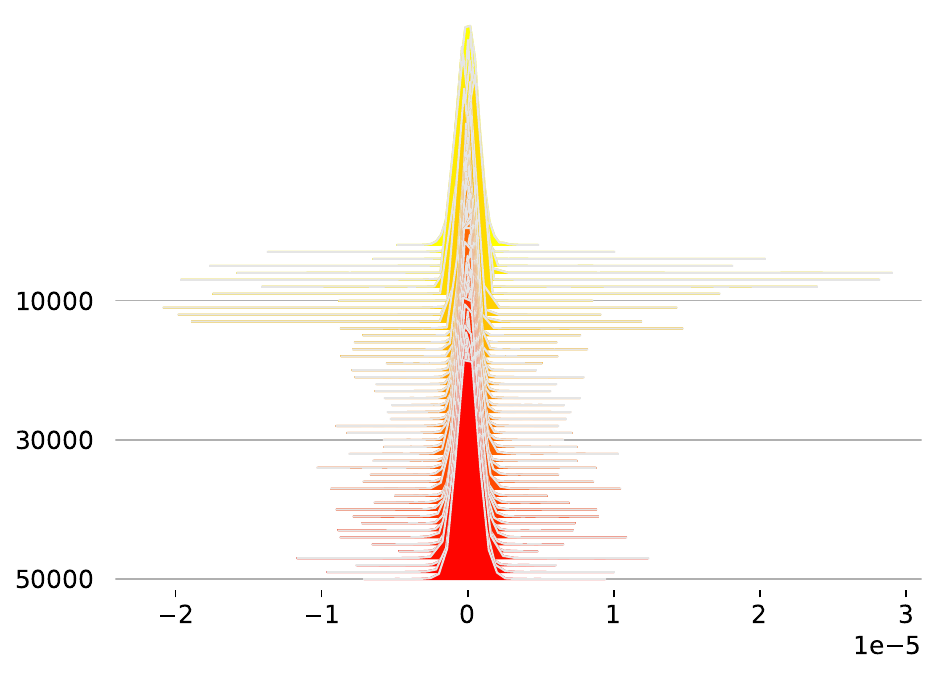}
            \includegraphics[width=0.42\textwidth]{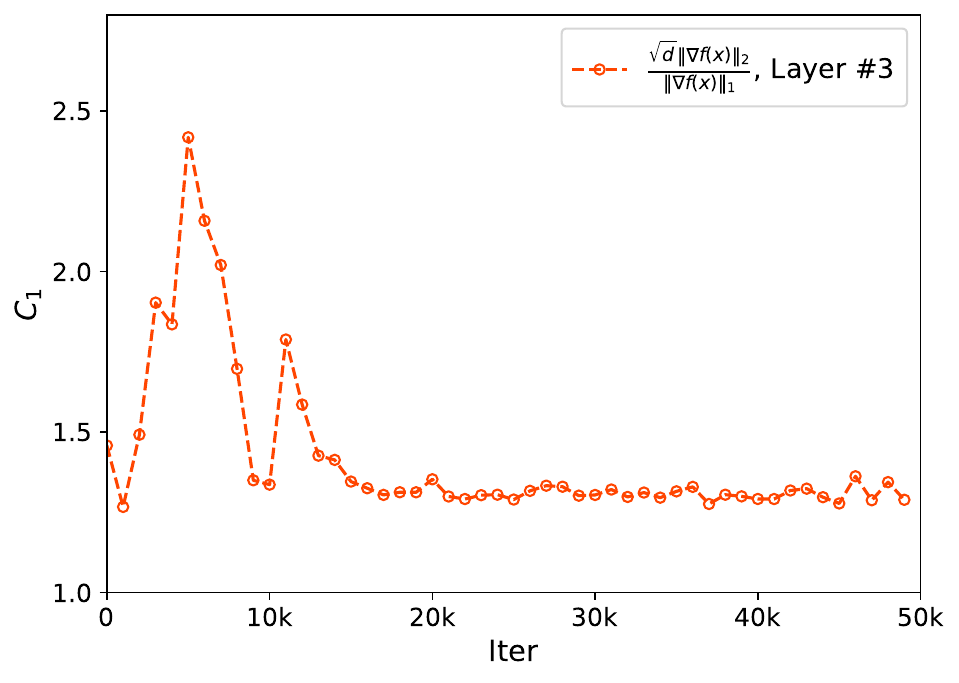}
		\end{minipage}
	}
  \vspace{-10pt}
  \caption{\small{The distribution and $C_1 = \frac{\sqrt{d}\Vert \nabla f(\bm{x}_t) \Vert_2}{\Vert \nabla f(\bm{x}_t) \Vert_1}$ for gradients across coordinates of (a) Layer\#32.conv.weight in ResNet-50 during training with Adam on ImageNet for 150 epochs, and (b) Layer\#3.self-attention.in-proj-weight in GPT-2 (350M) during training with Adam on OpenWebText for 50,000 iterations. Throughout training, $C_1$ remains consistently below 3, which is significantly smaller than $\sqrt{d}$, where $d$ represents the number of coordinates in the layers.} }
  \label{Figure.2}
\end{figure*}

 {Condition 3} commonly holds in practice, and we also empirically validated it in our experiments, as shown in Figure \ref{Figure.2}. Specifically, we employed Adam to train ResNet-50 on ImageNet and GPT-2 (350M) on OpenWebText. During training, we recorded the gradient $\nabla f(\bm{x}_t)$ for each coordinate in selected layers. Subsequently, we computed $ C_1 = \frac{\sqrt{d} \Vert \nabla f(\bm{x}_t) \Vert_2}{\Vert \nabla f(\bm{x}_t) \Vert_1} $. As shown in Figure \ref{Figure.2}, \( C_1 \) consistently remains below 3 throughout training, which is significantly smaller than \( \sqrt{d} \), where \( d \) represents the number of coordinates in the layers. This observation can be attributed to the fact that the coordinates of $ \nabla f(\bm{x}_t) $ tend to be densely clustered during training, as also depicted in Figure \ref{Figure.2}.

\section{Main Result}


In this section, we present our results on the convergence of Adam under two scenarios: 1). Assumption B.3 ($(L_0, L_1, q)$-smoothness) without reliance on Conditions 1-3. 2). Assumptions B.3 ($(L_0, L_1, q)$-smoothness) and C.3 ($p$-affine variance) with the inclusion of Conditions 1-3.

We first state the preliminary result  in Theorem \ref{lemma_convergence}.

\begin{theorem}
Let $\{x_t\}_{t=0}^{T-1}$ be generated by  Algorithm 1. Suppose that Assumptions A,  B.3,  and C.3, along with Condition 1,  hold. Define $u_t^{(j)} := \sfrac{\vert \bm{m}_t^{(j)} \vert}{(\sqrt{\bm{v}_t^{(j)}}+\epsilon)}$, $R:=\sfrac{(1-\beta_1)}{\sqrt{(1-\beta_2)(1-\frac{\beta_1^2}{\beta_2}})}$ , $\hat{L}:=L_0+(1-q)L_1$,  and $\hat{\sigma} := \sigma_0+ \sqrt{\frac{2-p}{2}}$. Choose $\beta_1 <\sqrt{\beta_2}$. Then,  it holds  for any $T \in {\mathbb N^{+}}$,
\begin{equation}
\small
\hspace{-5pt}
\begin{aligned}
& \frac{1}{T}\left(\sum_{t=0}^{T-1} {\mathbb E} [\Vert\bm{u}_t \circ \nabla F(\bm{x}_t) \Vert_1]- \left(\frac{\gamma R^2dqL_1}{2}+ {{2}C_0R\sqrt{d}\sigma_1\sqrt{p(1-\beta_1)}}+ \frac{2\gamma R^2{d}qL_1 }{1-\beta_1}\right)\sum_{t=0}^{T-1} \mathbb E [\Vert\nabla F(\bm{x}_t)\Vert_2]\right)  \\
&\le  \frac{ F(\bm{x}_{0}) - F^*}{\gamma T} +  \frac{2R\sqrt{d}\left \Vert \nabla F(\bm{x}_{0})\right\Vert_2}{T(1-\beta_1)} + {2\sqrt{1-\beta_1}R\sqrt{d}} \hat{\sigma}
+ \frac{2\gamma R^2 d \hat{L} }{1-\beta_1} + \frac{\gamma R^2{d}\hat{L}  }{2T}.
\end{aligned}
\label{conv_result}
\end{equation}
\label{lemma_convergence}
\vspace{-25pt}
\end{theorem}

When Assumption C.3 is replaced with Assumption C.1, Condition 1 becomes unnecessary, as detailed in the proof of Theorem \ref{lemma_convergence}. Building on the general result established in Theorem \ref{lemma_convergence}, we derive specific convergence bounds for the Adam algorithm under Assumptions A, B.3, and C.1, without requiring Conditions 1, 2, or 3.

\begin{corollary}[Convergence of Adam without Condition 1-3] Let $\{x_t\}_{t=0}^{T-1}$ be generated by  Algorithm 1. Suppose that  Assumptions A,  B.3 and C.1  hold. Define $u_t^{(j)} := \sfrac{\vert \bm{m}_t^{(j)} \vert}{(\sqrt{\bm{v}_t^{(j)}}+\epsilon)}$, $R:=\sfrac{1-\beta_1}{\sqrt{(1-\beta_2)(1-\sfrac{\beta_1^2}{\beta_2}})}$ and $\hat{L}:=L_0+(1-q)L_1$. Choose $\gamma=\frac{C_2}{T^{3/4}d^{\sfrac{1}{2}}}$, $\beta_1 < \sqrt{\beta_2}$, $1-\beta_1= \frac{C_3}{T^{\sfrac{1}{2}}}$ and $0< v \le \min_{t,j} u_{t}^{(j)} $. Then,  it holds  for any $T \in {\mathbb N^{+}}$ and $T \ge (\frac{4C_2 R^2d^{\sfrac{1}{2}}qL_1}{C_3v} +  (\frac{C_2 R^2d^{\sfrac{1}{2}}qL_1}{v})^{\sfrac{1}{3}})^4$,
\begin{equation}
\small
\begin{aligned}
\frac{1}{T}\sum_{t=0}^{T-1} \mathbb E [\Vert\nabla F(\bm{x}_t)\Vert_1] \le& \frac{1}{v}\left(\frac{2( F(\bm{x}_{0})- F(\bm{x}^*))}{C_2T^{\sfrac{1}{4}}d^{\sfrac{1}{2}}} + \frac{4Rd^{\sfrac{1}{2}}\left\Vert \nabla F(\bm{x}_{0})\right\Vert_2}{C_3T^{\sfrac{1}{2}}} \right. \\
+&\left.\frac{4C_3Rd^{\sfrac{1}{2}}\sigma_0}{T^{\sfrac{1}{4}}} +  \frac{4C_2 R^2d^{\sfrac{1}{2}}\hat{L}}{C_3T^{\sfrac{1}{4}}} + \frac{C_2R^2d^{\sfrac{1}{2}}\hat{L}}{T^{\sfrac{7}{4}}}\right).
\end{aligned}
\end{equation}
\label{theorem_convergence}
\vspace{-20pt}
\end{corollary}

Given the generic result in  Theorem \ref{lemma_convergence}, we derive the more comprehensive convergence bound of Adam under assumptions of Assumptions A, B.3 and C.3 with Condition 1, 2 and 3.

\begin{corollary} [Convergence of Adam with Condition 1-3]
Let $\{x_t\}_{t=0}^{T-1}$ be generated by  Algorithm 1. Suppose Assumptions A,  B.3 and C.3, along Conditions 1,2 and 3,  hold. Define $u_t := \sfrac{\vert \bm{m}_t^{(j)} \vert}{(\sqrt{\bm{v}_t^{(j)}}+\epsilon)}$, $R:=\sfrac{1-\beta_1}{\sqrt{(1-\beta_2)(1-\sfrac{\beta_1^2}{\beta_2}})}$, $\hat{L}:=L_0+(1-q)L_1$,  and $\hat{\sigma} := \sigma_0+ \sqrt{\frac{2-p}{2}}$.

\textbf{Case 1: General Setting with no Access to Oracles}

Choose  $\gamma=\frac{{C}_2}{T^{3/4}d^{\sfrac{1}{2}}}$, $\beta_1 <
  \sqrt{\beta_2}$, $1-\beta_1= \frac{C_3}{T^{\sfrac{1}{2}}}$,  and $0<\bar{v}\le \min_t \mathbb E [\bm{u}_t^{(j)}]  $.  Then, it holds for any $T \in {\mathbb N^{+}}$ and  $T \ge (\frac{4C_0C_1 {C}_3{\sfrac{1}{2}}p^{\sfrac{1}{2}}R\sigma_1}{\bar{v}}+\frac{4C_1{C_2} R^2qL_1}{{C}_3\bar{v}} +  (\frac{C_1{C}_2 R^2qL_1}{\bar{v}})^{\sfrac{1}{3}})^4$,
\begin{equation}
\small
\begin{aligned}
\frac{1}{T}\sum_{t=0}^{T-1} \mathbb E [\Vert\nabla F(\bm{x}_t)\Vert_2] \le& \frac{C_1}{\bar{v}}\left(\frac{2( F(\bm{x}_{0})- F(\bm{x}^*))}{C_2T^{\sfrac{1}{4}}d^{\sfrac{1}{2}}} + \frac{4R\left\Vert \nabla F(\bm{x}_{0})\right\Vert_2}{C_3T^{\sfrac{1}{2}}}
+ \frac{4C_3R\hat{\sigma}}{T^{\sfrac{1}{4}}} +  \frac{4C_2 R^2\hat{L}}{C_3T^{\sfrac{1}{4}}} + \frac{C_2R^2\hat{L}}{T^{\sfrac{7}{4}}} \right).
\end{aligned}
\end{equation}

\textbf{Case 2: Lowest-Bound Setting with Access to Oracles}
\vspace{5pt}

Choose  $\hat{C}_2=\frac{(F(\bm{x}_{0}) - F^*)^{\sfrac{3}{4}}}{2^{\sfrac{1}{4}}R\hat{\sigma}^{\sfrac{1}{2}}\hat{L}^{\sfrac{1}{4}}}$,  $\hat{C}_3=\frac{2^{\sfrac{1}{2}}\hat{L}^{\sfrac{1}{2}}(F(\bm{x}_{0}) - F^*)^{\sfrac{1}{2}}}{\hat{\sigma}}$, $\gamma=\frac{\hat{C}_2}{T^{3/4}d^{\sfrac{1}{2}}}$, $\beta_1 <
  \sqrt{\beta_2}$, $1-\beta_1= \frac{\hat{C}_3}{T^{\sfrac{1}{2}}}$, and  $0<\bar{v} \le \min_t \mathbb E [\bm{u}_t^{(j)}]  $.  Then,  for any $T \in {\mathbb N^{+}}$ and $T \ge (\frac{4C_0C_1 \hat{C}_3{\sfrac{1}{2}}p^{\sfrac{1}{2}}R\sigma_1}{\bar{v}}+\frac{4C_1\hat{C_2} R^2qL_1}{\hat{C}_3\bar{v}} +  (\frac{C_1\hat{C}_2 R^2qL_1}{\bar{v}})^{\sfrac{1}{3}})^4$ the right-hand side of Eq. (\ref{conv_result}) reaches the lowest bound, \emph{i.e.},
\begin{equation}
\small
\begin{aligned}
\frac{1}{T}\sum_{t=0}^{T-1} \mathbb E [\Vert\nabla F(\bm{x}_t)\Vert_2] \le& \frac{C_1} {\bar{v}} \left( \frac{512^{\sfrac{1}{4}} R\hat{\sigma}^{\sfrac{1}{2}}\hat{L}^{\sfrac{1}{4}}(F(\bm{x}_{0}) - F^*)}{T^{\sfrac{1}{4}}} + \frac{2C_1\hat{C}_3R\left\Vert \nabla F(\bm{x}_{0})\right\Vert_2}{T^{\sfrac{3}{4}}} +  \frac{  \hat{C_2}  R^2\hat{L}  }{T^{\sfrac{7}{4}}}\right).
\end{aligned}
\end{equation}
\label{theorem_convergence_1}
\vspace{-20pt}
\end{corollary}

We have some observations from Theorem \ref{lemma_convergence}, Corollary \ref{theorem_convergence}  and Corollary \ref{theorem_convergence_1} below.

\textbf{Finding 1.}                                                                                                                                                                  To the best of our knowledge, we are the first to formally analyze Adam under the weak assumptions of generalized non-uniform $(L_0, L_1, q)$-smoothness (Assumption B.3) and $p$-affine variance (Assumption C.3). We also prove that $\frac{1}{T}\sum_{t=0}^{T-1} \mathbb E [\Vert\nabla F(\bm{x}_t)\Vert_2]$ for Adam achieves a tighter bound of {\small ${\cal O} \left(\frac{1}{T^{\sfrac{1}{4}}}\right)$}, compared to the previous {\small ${\cal O} \left(\frac{\ln T}{T^{\sfrac{1}{4}}}\right)$} \citep{Adam_type_proof_2018,Simple_Adam_proof_2020,Prac_Adam_proof_2022, Relaxed_Adam_proof_2024}. Furthermore, earlier works demonstrated that Adam's convergence rates were  dependent on the model dimensionality $d$ and the numerical-stability $\epsilon$ \citep{Adam_type_proof_2018,Simple_Adam_proof_2020,Prac_Adam_proof_2022, Relaxed_Adam_proof_2024}, which makes them unsuitable to analyze large-scale LLM training. However, as shown in Corollary \ref{theorem_convergence_1}, we prove that Adam achieves dimension-free and $\epsilon$-free convergence, similar to SGD \citep{Opt_Review_2018}. Notably, previous studies required a learning rate of $\gamma = 1 - O(\frac{1}{T})$ to reach the optimal convergence rate of {\small ${\cal O} \left(\frac{\ln T}{T^{\sfrac{1}{4}}}\right)$}. This causes $\bm{v}_t$ to closely resemble a plain average of the past $T$ squared gradients, reducing Adam almost to AdaGrad \citep{Simple_Adam_proof_2020}. By contrast, we only require the learning rate to satisfy $\gamma = 1 - O(\frac{1}{T^{\sfrac{3}{4}}})$, which better aligns with Adam's original design.

\textbf{Finding 2.} In the proof of Theorem \ref{lemma_convergence}, Condition 1 is no longer necessary when Assumption C.1 is used. As demonstrated in Corollary \ref{theorem_convergence}, even in the absence of Conditions 1, 2, and 3, and assuming no access to the oracle values $L_0$, $L_1$, $\sigma_0$, and $\sigma_1$, we can still theoretically establish that $\frac{1}{T} \sum_{t=0}^{T-1} \mathbb{E} [\Vert \nabla F(\bm{x}_t) \Vert_2]$ for Adam converges at the rate of $O\left(\frac{1}{T^{\sfrac{1}{4}}}\right)$, under the assumptions of  $(L_0, L_1, q)$-smoothness and  affine variance.

\textbf{Finding 3.} The momentum coefficients $\beta_1$ and $\beta_2$ play a crucial role in Adam's convergence. As shown in Theorem \ref{lemma_convergence}, when $\beta_1 = \beta_2 = 0$, Adam reduces to signSGD, which converges at best to a bounded region where $\frac{1}{T}\sum_{t=0}^{T-1} \mathbb E [\Vert\nabla F(\bm{x}_t)\Vert_2] \le O(\frac{1}{T^{\sfrac{1}{4}}} + \sigma_0)$. This result aligns with prior work \citep{signSGD_2018}. In contrast, sufficiently large values of $\beta_1$ and $\beta_2$ ensure Adam achieves the convergence rate of $O(\frac{1}{T^{\sfrac{1}{4}}})$. For SGD, however, momentum has minimal impact on the optimal theoretical convergence rate, as both momentum-SGD and vanilla SGD converge at $O(\frac{1}{T^{\sfrac{1}{4}}})$ \citep{Opt_Review_2018,SGDM_convergence_2020}.

\textbf{Finding 4.}  Case 2 in Corollary \ref{theorem_convergence_1} states that Adam's learning rate must satisfy $\gamma = O(\frac{1}{\sqrt{d}})$ to achieve the optimal convergence rate. This implies that, with fixed other hyperparameters, larger model sizes require smaller optimal learning rates. This observation has been empirically validated by practitioners training the Llama family of models across different sizes \citep{Llama2023,llama2_2023,llama3_2024}. Interestingly, a similar theoretical conclusion appears in the work on Maximal Update Parametrization ($\mu P$) \citep{MuP_2022}.

\section{Proof Sketch}

In this section, we present the core ideas underlying the convergence proofs for Theorem \ref{lemma_convergence} and Corollary \ref{theorem_convergence_1}. The proof ideas for Corollary \ref{theorem_convergence} are similar, and thus, we omit the details for simplicity.

Our main contribution lies in opening up  a new approach to proving the convergence of Adam. All existing theoretical convergence proofs follow a path-dependent approach, treating Adam as a preconditioned SGD with momentum, as originally presented in the Adam paper \citep{Adam2015}. Specifically, the update rule is defined as: \emph{i.e.},$\bm{x}_t+1 = \bm{x}_t - \frac{\gamma_t}{{\sqrt{\bm{v}_t}+\epsilon}} \circ \bm{m}_t$,  where ${\sqrt{\bm{v}_t} + \epsilon}$ is used to precondition $\bm{m}_t$, and the effective learning rate is $\frac{\gamma_t}{\sqrt{\bm{v}_t} + \epsilon}$. This approach, however, encounters two intractable issues: ($i$) the effective learning rate $\frac{\gamma_t}{\sqrt{\bm{v}_t} + \epsilon}$ is not necessarily monotone-decreasing, and ($ii$) the random variable $\bm{v}_t$ is not independent of $\bm{g}_t$ or $\bm{m}_t$. To address these challenges, the proofs in previous works became complicated, lengthy, and opaque, making them difficult to verify and extend. In contrast, we treat Adam as a whole stochastic sign-like descent algorithm, \emph{i.e.}, $\bm{x}_{t+1} = \bm{x}_t - \gamma_t\frac{|\bm{m}_t|}{{\sqrt{\bm{v}_t}+\epsilon}} \circ {\rm Sign}(\bm{m}_t)$ where we consider the term $\frac{|\bm{m}_t|}{\sqrt{\bm{v}_t} + \epsilon}$ as a single random variable. This transformation not only circumvents the problems mentioned above but also simplifies the proof process. We now provide a sketch of the proof.

Under Assumption B.3, we obtain (details please refer to Lamma 1 in the Appendix):
\begin{equation}
\small
{\mathbb E}[F(\bm{x}_{t+1})] \le  F(\bm{x}_t) + {\mathbb E}[\langle \nabla F(\bm{x}_t), \bm{x}_{t+1} -\bm{x}_t \rangle] + \frac{L_0 + L_1\Vert \nabla F(\bm{x}_t) \Vert_2^q}{2}{\mathbb E}[\Vert \bm{x}_{t+1} - \bm{x}_t\Vert_2^2].
\end{equation}

Defining $\bm{u}_t := \frac{\vert \bm{m}_t \vert}{\sqrt{\bm{v}_t}+\epsilon}$,  the update rule becomes $\bm{x}_{t+1} = \bm{x}_t - \gamma \frac{\bm{m}_t}{\sqrt{\bm{v}_t}+\epsilon} = \bm{x}_t - \gamma \frac{\vert \bm{m}_t \vert}{\sqrt{\bm{v}_t}+\epsilon} \circ \frac{\bm{m}_t}{|\bm{m}_t|} = \bm{x}_t - \gamma \bm{u}_t \circ {\rm Sign}(\bm{m}_t) $. We further have
\begin{equation}
\small
\begin{aligned}
{\mathbb E}[F(\bm{x}_{t+1})] \le & F(\bm{x}_t) - \gamma {\mathbb E}[\Vert \bm{u}_t \nabla  F(\bm{x}_t) \Vert_1] + \underbrace{\gamma {\mathbb E}[\left\langle \nabla F(\bm{x}_t),  \bm{u}_t \circ ({\rm Sign}(\nabla F(\bm{x}_t)) - {\rm Sign}(\bm{m}_t) )\right\rangle]}_{{\cal T}_1}  \\
&+\underbrace{\frac{\gamma^2(L_0 + L_q\Vert \nabla F(\bm{x}_t) \Vert_2^q)}{2}{\mathbb E}[\Vert \bm{u}_t \Vert_2^2]}_{{\cal T}_2}.
\end{aligned}
\end{equation}

Next, we define $R:=\frac{1-\beta_1}{\sqrt{(1-\beta_2)(1-\sfrac{\beta_1^2}{\beta_2}})}$, and by applying Lemma 2 in the appendix, we obtain that $\bm{u}_t^{(j)} \le R$. Furthermore, Lemma 3 in the appendix indicates ${\mathbb E}[\vert{\rm Sign}(\nabla F(\bm{x}_t^{(j)})) - {\rm Sign}(\bm{m}_t^{(j)})\vert] \le 2\frac{{\mathbb E}[\vert \nabla F(\bm{x}_t^{(j)}) - \bm{m}_t^{(j)}\vert ]}{\vert\nabla F(\bm{x}_t^{(j)})\vert}$, which leads to ${\cal T}_1 \le 2\gamma R\sqrt{d}{\mathbb E}[\Vert \nabla F(\bm{x}_t) - \bm{m}_t\Vert_2 ]$.

Employing the bound $\bm{u}_t^{(j)} \le R$ above and applying Young's inequality, we obtain ${\cal T}_2 \le \frac{\gamma^2 R^2d(L_0 + L_1((1-q)+q{\mathbb E}[\Vert \nabla F(\bm{x}_t) \Vert_2]))}{2}$.

By taking the expectation over the first to the $(T-1)$-th iteration, and then summing and rearranging the terms, we obtain:
\begin{equation}
\small
\begin{aligned}
&\frac{1}{T}\sum_{t=0}^{T-1}{\mathbb E} [\Vert\bm{u}_t \circ \nabla F(\bm{x}_t) \Vert_1]- \frac{\gamma R^2qL_1{d}}{2T}\sum_{t=0}^{T-1} \mathbb E [\Vert\nabla F(\bm{x}_t)\Vert_2] \\
\le & \frac{ F(\bm{x}_{0}) - F^*}{\gamma T}
 + \frac{2R\sqrt{d}}{T}\sum_{t=0}^{T-1} {\mathbb E} [\Vert \bm{m}_t - \nabla F(\bm{x}_t)\Vert_2]
+ \frac{\gamma R^2{d}(L_0+(1-q)L_1)  }{2T}.
\end{aligned}
\end{equation}

We now divide-and-conquer prove that
\begin{equation}
\small
\begin{aligned}
\frac{1}{T}\sum_{t=1}^{T}{\mathbb E}\left[\Vert \bm{m}_t - \nabla F(\bm{x}_t) \Vert_2\right]  \le & \frac{\left\Vert \nabla F(\bm{x}_{0})\right\Vert_2}{T(1-\beta_1)}+ {\sqrt{1-\beta_1}}\left(\sigma_0 + \sqrt{\frac{2-p}{2}} \sigma_1\right) \\
+& {C_0\sigma_1\sqrt{p(1-\beta_1)}} \cdot\frac{1}{T}\sum_{t=0}^{T-1} \Vert \nabla F(x_t) \Vert_2 \\
 +& \frac{\gamma R \sqrt{d} (L_0+(1-q)L_1) }{1-\beta_1} + \frac{\gamma R \sqrt{d} qL_1}{1-\beta_1}\cdot \frac{1}{T}\sum_{t=0}^{T-1}{\mathbb E}[\Vert\nabla F(\bm{x}_{t})\Vert_2]\\
\end{aligned}
\end{equation}

Then, this leads to the conclusion in Theorem  \ref{lemma_convergence}, \emph{i.e.},

\begin{equation}
\small
\begin{aligned}
& \frac{1}{T}\left(\sum_{t=0}^{T-1} {\mathbb E} [\Vert\bm{u}_t \circ \nabla F(\bm{x}_t) \Vert_1]- \left(\frac{\gamma R^2dqL_1}{2}+ {{2}C_0R\sqrt{d}\sigma_1\sqrt{p(1-\beta_1)}}+ \frac{2\gamma R^2{d}qL_1 }{1-\beta_1}\right)\sum_{t=0}^{T-1} \mathbb E [\Vert\nabla F(\bm{x}_t)\Vert_2]\right)  \\
&\le  \frac{ F(\bm{x}_{0}) - F^*}{\gamma T} +  \frac{2R\sqrt{d}\left \Vert \nabla F(\bm{x}_{0})\right\Vert_2}{T(1-\beta_1)} + {2\sqrt{1-\beta_1}R\sqrt{d}} \hat{\sigma}
+ \frac{2\gamma R^2 d \hat{L} }{1-\beta_1} + \frac{\gamma R^2{d}\hat{L}  }{2T}.
\end{aligned}
\end{equation}
where $\hat{L}:=L_0+(1-q)L_1$,  and $\hat{\sigma} := \sigma_0+ \sqrt{\frac{2-p}{2}}$.

By choosing $\bar{v}= \min_t \mathbb E [\bm{u}_t^{(j)}] $  and applying Condition 2 and Condition 3,  we obtain
\begin{equation}
\small
\sum_{t=0}^{T-1} {\mathbb E} [\Vert\bm{u}_t \circ \nabla F(\bm{x}_t) \Vert_1] =  \sum_{t=0}^{T-1} {\mathbb E} [\bm{u}_t^{(j)}] {\mathbb E} [ \Vert \nabla F(\bm{x}_t) \Vert_1] \ge \bar{v}\sum_{t=0}^{T-1}{\mathbb E} [ \Vert (\nabla F(\bm{x}_t)) \Vert_1] = \frac{\bar{v}\sqrt{d}}{C_1}\sum_{t=0}^{T-1} {\mathbb E} [ \Vert (\nabla F(\bm{x}_t)) \Vert_2].
\end{equation}

Next, by choosing  $\gamma=\frac{C_2}{T^{\sfrac{3}{4}}d^{\sfrac{1}{2}}}$, $1-\beta_1= \frac{C_3}{T^{\sfrac{1}{2}}}$ and setting $T \ge (\frac{4C_2 R^2qL_1}{C_3\bar{v}} +  {4}C_0C_1\sqrt{C_3}R\sigma_1\sqrt{p} +  (\frac{C_2 R^2qL_1}{\bar{v}})^{\sfrac{1}{3}})^4$, applying following Lemma 4 in the appendix, we have
\begin{equation}
\small
\frac{\gamma C_1R^2\sqrt{d}qL_1}{2} + {{2}C_0C_1R\sigma_1\sqrt{p(1-\beta_1)}} + \frac{2\gamma C_1 R^2\sqrt{d}qL_1 }{1-\beta_1} \le \frac{\bar{v}}{2}.
\end{equation}

Then, we arrive Case 1 of Corollary \ref{theorem_convergence_1}, \emph{i.e.},
\begin{equation}
\small
\begin{aligned}
\frac{1}{T}\sum_{t=0}^{T-1} \mathbb E [\Vert\nabla F(\bm{x}_t)\Vert_2] \le& \frac{C_1}{\bar{v}}\left(\frac{2L_1( F(\bm{x}_{0})- F(\bm{x}^*))}{C_2T^{\sfrac{1}{4}}d^{\sfrac{1}{2}}} + \frac{4R\left\Vert \nabla F(\bm{x}_{0})\right\Vert_2}{C_3T^{\sfrac{1}{2}}}
+ \frac{4C_3R\hat{\sigma}}{T^{\sfrac{1}{4}}} +  \frac{4C_2 R^2\hat{L}}{C_3T^{\sfrac{1}{4}}} + \frac{C_2R^2\hat{L}}{T^{\sfrac{7}{4}}} \right).
\end{aligned}
\end{equation}

Using generalized Young's inequality, we minimize the bottleneck terms to achieve the lowest bound on the right-hand side of Eq. (\ref{conv_result}) in Theorem \ref{lemma_convergence}, \emph{i.e.},
\begin{equation}
\small
\begin{aligned}
& \frac{ C_1(F(\bm{x}_{0}) - F^*)}{\gamma T \sqrt{d}} + {2C_1\sqrt{1-\beta_1}R}\hat{\sigma}
+ \frac{2C_1\gamma R^2 \sqrt{d} \hat{L} }{1-\beta_1}
\ge \frac{512^{\sfrac{1}{4}} C_1R\hat{\sigma}^{\sfrac{1}{2}}\hat{L}^{\sfrac{1}{4}}(F(\bm{x}_{0}) - F^*)}{T^{\sfrac{1}{4}}},
\end{aligned}
\end{equation}
where the lowest bound achieved if and only if $\gamma = \frac{(F(\bm{x}_{0}) - F^*)^{\sfrac{3}{4}}}{2^{\sfrac{1}{4}}T^{\sfrac{3}{4}}d^{\sfrac{1}{2}}R\hat{\sigma}^{\sfrac{1}{2}}\hat{L}^{\sfrac{1}{4}}}$ and $1 - \beta_1 = \frac{2^{\sfrac{1}{2}}\hat{L}^{\sfrac{1}{2}}(F(\bm{x}_{0}) - F^*)^{\sfrac{1}{2}}}{T^{\sfrac{1}{2}}\hat{\sigma}}$.

Now, choosing  $\hat{C}_2=\frac{(F(\bm{x}_{0}) - F^*)^{\sfrac{3}{4}}}{2^{\sfrac{1}{4}}R\hat{\sigma}^{\sfrac{1}{2}}\hat{L}^{\sfrac{1}{4}}}$,  $\hat{C}_3=\frac{2^{\sfrac{1}{2}}\hat{L}^{\sfrac{1}{2}}(F(\bm{x}_{0}) - F^*)^{\sfrac{1}{2}}}{\hat{\sigma}}$, $\gamma=\frac{\hat{C}_2}{T^{3/4}d^{\sfrac{1}{2}}}$, $1-\beta_1= \frac{\hat{C}_3}{T^{\sfrac{1}{2}}}$ and setting $T \ge (\frac{4C_0C_1 \hat{C}_3{\sfrac{1}{2}}p^{\sfrac{1}{2}}R\sigma_1}{\bar{v}}+\frac{4C_1\hat{C_2} R^2qL_1}{\hat{C}_3\bar{v}} +  (\frac{C_1\hat{C}_2 R^2qL_1}{\bar{v}})^{\sfrac{1}{3}})^4$, we arrive Case 2 of Corollary \ref{theorem_convergence_1}, \emph{i.e.},
\begin{equation}
\small
\begin{aligned}
\frac{1}{T}\sum_{t=0}^{T-1} \mathbb E [\Vert\nabla F(\bm{x}_t)\Vert_2] \le& \frac{C_1} {\bar{v}} \left( \frac{512^{\sfrac{1}{4}} R\hat{\sigma}^{\sfrac{1}{2}}\hat{L}^{\sfrac{1}{4}}(F(\bm{x}_{0}) - F^*)}{T^{\sfrac{1}{4}}} + \frac{2C_1\hat{C}_3R\left\Vert \nabla F(\bm{x}_{0})\right\Vert_2}{T^{\sfrac{3}{4}}} +  \frac{  \hat{C_2}  R^2\hat{L}  }{T^{\sfrac{7}{4}}}\right).
\end{aligned}
\end{equation}

\section{Conclusion}

This work breaks with convention and provides a pioneering reinterpretation of Adam as a sign-like descent algorithm to analyze the convergence, simplifying its theoretical analysis and addressing limitations of the traditional preconditioned perspective. By treating $\frac{|\bm{m}_t|}{{\sqrt{\bm{v}_t}+\epsilon}}$ as a unified random variable, It is  the first time to proved that Adam dimension-freely and $\epsilon$-freely achieves the optimal convergence rate of $\mathcal{O}(\frac{1}{T^{\sfrac{1}{4}}})$ under the weak assumptions of generalized $(L_0, L_1,q)$-smoothness and affine variance.  Additionally, we provide actionable insights for hyperparameter tuning, bridging theory and practice. This work advances the theoretical convergence analysis of adaptive optimizers and offers a foundation for further research and development.

%
%

\bibliography{./refs/refs}
\bibliographystyle{plain}


\newpage
\appendix

\begin{center}
\large
\textbf{Appendix}
\end{center}

\textbf{Limitations.}  While the proposed novel route for the convergence proof of Adam simplifies the theoretical analysis, extending the proof to weaker assumptions (Assumptions C.3) compared to prior work needs the introduction of additional constraints (Conditions 1–3), although these conditions are commonly satisfied in practical settings that are demonstrated in Section 3.

\textbf{Broader Impacts.}  This paper advances the theoretical understanding of Adam's convergence properties, which may contribute to more efficient and reliable training of deep neural networks. The main contributions of this paper are theoretical, and thus it do not inherently raise potential ethic and societal concerns.

\section{Relate Work}
There is a large amount of works on the theoretical analysis of stochastic descents algorithms. In this section, we list the most related references and make comparison with our work.

\textbf{Convergence with Weak Assumptions.} \cite{SGD_affine_var_2000} first theoretically analyze  SGD under the assumption of affine variances, obtaining an asymptotic convergence result. Until 2018, \cite{Opt_Review_2018} proved the non-asymptotic convergence rate of { \small $\Vert \nabla F(x) \Vert_2$ for SGD up to ${\cal O} \left(\frac{{\rm poly ( ln}T)}{T^{\sfrac{1}{4}}}\right)$}, which matched its provable rate with the bounded variance condition.  In terms of adaptive optimizers, \cite{AdaGrad-Norm_affine_var_2022} investigated the convergence rate of AdaGrad-Norm with the affine variance, and proved the rate could achieve {\small ${\cal O} \left(\frac{{\rm poly ( ln}T)}{T^{\sfrac{1}{4}}}\right)$} as well when{\small $\sigma_1 = {\cal O} \left(\frac{1}{\sqrt{T}}\right)$}. \cite{AdaGrad_affine_var_2023}  proved the AdaGrad-Norm obtained a similar convergence rate with no restriction over $\sigma_1$, and it further demonstrate  vanilla AdaGrad could also achieve the same convergence rate  under a stronger assumption of coordinate-wise affine variances. Meanwhile, \cite{AdaGrad-norm_prob_affine_var_2023} provided a probabilistic convergence rate for AdaGrad-Norm. Noted that \cite{RMSProp_affine_var_2021} and \cite{Adam_proof_2022} respectively proved random-shuffled AMSProp and Adam will converge to the neighbourhood of stationary points  with the rate {\small ${\cal O} \left(\frac{{\rm poly ( ln}E)}{E^{\sfrac{1}{4}}} + \sigma_0 \right)$} where $E$ is the number of  epoches rather than iterations  under the affine growth condition that is equivalent to the affine variance condition, which is much slower than the optimal rate. Recently, \cite{Adam_affine_var_2023} provably demonstrate the convergence rate of vanilla Adam in high probability perspective, but it only works with the stronger coordinate-wise affine variance.

\cite{Clipped_SGD_Proof_2019} first introduced the unciform $(L_0, L_1)$-smooth condition to theoretically explain why Clipped-SGD converges faster than vanilla SGD, and they also empirically verified that local smoothness indeed varies with the norm of gradients during DNN training. \cite{Imprv_clipping_SGD_2020} posits that it is also equivalent to an affine form of the gradient norm for the first-order differentiable function. Then, the this relaxed assumption is extended to analyzing clipping-SGD with momentum \cite{Imprv_clipping_SGD_2020},  distributionally robust optimization \cite{DRO_gene_smooth_2021}, normalized SGD with momentum \cite{NSGDM_gene_smooth_2024}, generalized signSGD \cite{Generalized_signSGD_2022}.  Recently, \cite{Adaptivity_Adam_proof_2024} theoretically analyzing random-shuffled Adam under this condition, but its convergence rate is provable {\small ${\cal O} \left(\frac{{\rm poly ( ln}E)}{E^{\sfrac{1}{4}}} \right)$} where $E$ is the number of epoch, just like \cite{Adam_proof_2022}.
\cite{Relaxed_Adam_proof_2024} further extended the linear $(L_0, L_1)$-smooth to the generalized polynomial version, and proved that Adam will converged to {\small ${\cal O} \left(\frac{{\rm poly ( ln}T)}{T^{\sfrac{1}{4}}}\right)$} with the weaker assumption. However, the bound heavily relies on a large $\epsilon$ , but Adam with a large $\epsilon$ is essentially similar to SGD and loses the nature of adaptivity.

Furthermore, \cite{AdaGrad_weak_assum_2023, AdaGrad_weak_assum_2023_1}  theoretically  analyze  AdaGrad under the weak assumptions of both  affine variance and  unciform $(L_0, L_1)$-smooth, also obtaining the rate of {\small ${\cal O} \left(\frac{{\rm poly ( ln}T)}{T^{\sfrac{1}{4}}}\right)$}. \cite{Separa_Adam_proof_2024} also provable provide a high probability convergence  of a simplified Adam-Norm with the both weak assumptions. More recently, \cite{Adam_weak_assum_2025} prove the convergence rate of vanilla Adam with both the affine variance condition and  the $(L_0, L_1)$-smooth condition, but its proof have a fatal error that may vacuum the validity (The constant and $T$-independent  $G$  is the premise of the theoretical analysis, but $1-\beta_2 = {\cal O}(\frac{1}{T})$, $1-\beta_2 = \frac{c}{T}$, Eq. (41), Eq. (56) and Eq. (58) in the proof suggest that $\Vert\bar{\bm{g}}_{t+1}\Vert^2$ should be larger than ${\cal O}(T)$. ).

In comparison to the convergence analyses above,  we are the first to formally analyze vanilla  under the weak assumptions of the generalized affine variance and  the generalized unciform $(L_0, L_1)$-smooth, achieving a tighter bound of {\small ${\cal O} \left(\frac{1}{T^{\sfrac{1}{4}}}\right)$} rather than{\small ${\cal O} \left(\frac{{\rm poly ( ln}T)}{T^{\sfrac{1}{4}}}\right)$}.

\textbf{Convergence of Sign Descent.} Sign-based algorithms that simply exploits the signs of gradients could date back to RPROP (\cite{RPROP_1993}). \cite{1-bit-SGD_2014,1-bit-SGD_2015} proposed 1-bit SGD and empirically demonstrate it achieve good performance while dramatically reducing the communication costs in distributed system. The non-stochastic convergence proof of signSGD was first analyzed in (\cite{signSGD_proof_2016}) under the Polyak-{\L}ojasiewicz condition. Then,  \cite{signSGD_2018} systematically establish the convergence rate of signSGD in stochastic and non-convex scenario, but it required an increasing batch size up to $O(\sqrt{n})$ where $n$ is the number of samples to guarantee convergence. Then, \cite{signSGD_proof_2023} first proved that momentum can ensures the convergence of signSGD without increasing batch size. \cite{Lion2023} employs an AutoML method to discover an effective optimizer, { Lion}, resembling  signSGD with momentum, and demonstrate superior performance to Adam across diverse DNN models. \cite{Lion_explain_2023}  theoretically analyzed the efficacy of Lion but did not provide the convergence proof.  Meanwhile, the original version of Adam, RMSProp \cite{RMSProp2012}, were developed from the sign-based Rprop.  \cite{MSSD_2018} also found that sign descent algorithms has a deep connection with Adam. Recently, \cite{Explian_Adam_SGD_2023,Explian_Adam_SGD_2024} empirically showcase that the sign-like property of Adam is just the primary reason behind its superior performance for training DNNs. In short, signSGD has a close connection with Adam, but the existing convergence proofs of Adam were not built upon this connection.

To the best of our knowledge, our work is the first to prove the convergence rate of Adam from the perspective of sign descent, and the proof, thereby, becomes considerably simple, compared to the previous theoretical proofs of Adam.

\section{Theoretical Analysis}
\subsection {Proof of Proposition 2.1}


\textbf{Proof.}  It is known that

 \begin{equation}
 \begin{aligned}
 \frac{1}{T}\sum_{t=0}^{T-1} \Vert \nabla F(\bm{x}_t) \Vert_2^2 = & {\cal O } \left( \frac{1}{T}\sum_{t=0}^{T-1} \frac{1}{t^{2\alpha}}\right) \\
 \le & {\cal O } \left(\frac{1}{T}\int_{t=0}^{T-1}  \frac{1}{t^{2\alpha}} dt \right) \\
  = & {\cal O } \left( \frac{T^{1-2\alpha} -1} {T(1-2\alpha)} \right)  \\
 \le & {\cal O } \left( \frac{c^2} {1-2\alpha} \cdot \frac{1}{T^{2\alpha}}\right).
 \end{aligned}
 \end{equation}

 and

 \begin{equation}
 \begin{aligned}
\left(\frac{1}{T}\sum_{t=0}^{T-1} \Vert \nabla F(\bm{x}_t)\Vert_2\right)^2 = & {\cal O } \left( \left(\frac{1}{T}\sum_{t=0}^{T-1} \frac{1}{t^{\alpha}}\right)^2 \right) \\
\ge & {\cal O } \left(\left(\frac{1}{T}\int_{t=2}^{T+1} \frac{1}{t^{\alpha}} dt\right)^2\right) \\
\ge & {\cal O } \left(\frac{1}{(1-\alpha)^2}\cdot\left(\frac{1}{T^{2\alpha}} - \frac{2^{1-\alpha}}{T^{1+\alpha}}\right)\right).
\end{aligned}
 \end{equation}

It is easy to verify that when $T \ge {2^{\frac{2-\alpha}{1-\alpha}}}$, $\frac{2^{1-\alpha}}{T^{1+\alpha}}\le \frac{1}{2}\cdot\frac{1}{T^{2\alpha}}$. Moreover, we can obtain ${2^{\frac{2-\alpha}{1-\alpha}}}\le 8$ due to $0\le \alpha < \frac{1}{2}$. Hence, when $T \ge 8$, we have

\begin{equation}
 \frac{\frac{1}{T}\sum_{t=0}^{T-1} \Vert \nabla F(\bm{x}_t) \Vert_2^2}{ \left(\frac{1}{T}\sum_{t=0}^{T-1} \Vert \nabla F(\bm{x}_t)\Vert_2\right)^2} \le {\cal O } \left( \frac{2(1-\alpha)^2}{1-2\alpha} \right).
\end{equation}

\subsection{Useful Lemmas}

\begin{lemma}
Under Assumption B.3, for any $\bm{x}, \bm{y} \in {\mathbb R}^d $, the function obeys

\begin{equation}
F(\bm{y}) \le  F(\bm{x}) + \langle \nabla F(\bm{x}), \bm{y} -\bm{x} \rangle + \frac{L_0 + L_1\Vert \nabla F(\bm{x}) \Vert_2^q}{2}\Vert \bm{y} - \bm{x}\Vert_2^2.
\end{equation}
\label{lamma_smooth}
\end{lemma}
\textbf{Proof.}  For any $\bm{x}, \bm{y} \in {\mathbb R}^d $, we have
\begin{equation}
\begin{aligned}
F(\bm{y}) = & F(\bm{x}) + \int_{0}^1 \langle \nabla F(\bm{x} + t(\bm{y}-\bm{x})), \bm{y}-\bm{x} \rangle dt \\
= & F(\bm{x}) + \langle \nabla F(\bm{x}), \bm{y}-\bm{x} \rangle + \int_{0}^1 \langle \nabla F(\bm{x} + t(\bm{y}-\bm{x}))-\nabla F(\bm{x}), \bm{y}-\bm{x} \rangle dt \\
\overset{(i)}{\le} & F(\bm{x}) + \langle \nabla F(\bm{x}), \bm{y}-\bm{x} \rangle + \int_0^1 \Vert \nabla F(\bm{x} + t(\bm{y}-\bm{x}))-\nabla F(\bm{x})\Vert_2 \Vert\bm{y}-\bm{x}\Vert_2 dt \\
\overset{(ii)}{\le} & F(\bm{x}) + \langle \nabla F(\bm{x}), \bm{y}-\bm{x} \rangle + (L_0 + L_1 \Vert \nabla  F(\bm{x}) \Vert_2^q)\Vert\bm{y}-\bm{x}\Vert_2^2 \int_0^1 t d_t \\
=&  F(\bm{x}) + \langle \nabla F(\bm{x}), \bm{y} -\bm{x} \rangle + \frac{L_0 + L_1\Vert \nabla F(\bm{x}) \Vert_2^q}{2}\Vert \bm{y} - \bm{x}\Vert_2^2,
\end{aligned}
\end{equation}
where$(i)$ holds due to Cauchy-Schwarz inequality, and  $(ii)$ holds due to Assumption 2.

\begin{lemma}
 Let the sequences $\{\bm{m}_t\}$ and $ \{\bm{v}_t\}$ be generated by Adam in Algorithm 1 with omitted bias correction. If the moving average coefficients $\beta_1, \beta_2$ are constant and satisfy $\beta_1^2 <
  \beta_2$, then

  (1) For any $j \in [d]$,  it holds that
  \begin{equation}
  \frac{\vert\bm{m}_t^{(j)}\vert}{\sqrt{\bm{v}_t^{(j)}}+\epsilon} \le  \frac{1-\beta_1}{\sqrt{1-\beta_2}\sqrt{1-\frac{\beta_1^2}{\beta_2}}}.
  \end{equation}

  (2) The maximal value of $\frac{1-\beta_1}{\sqrt{1-\beta_2}\sqrt{1-\frac{\beta_1^2}{\beta_2}}}$ is lower bounded by $1$, and  upper bounded by $\sqrt{\frac{1-\beta_1}{1-\beta_2}}$ when $\beta_1 \le \beta_2$.

  \label{lemma.1}
  \end{lemma}

\textbf{Proof.} (1) Recalling Adam, we know
  \begin{equation}
  \begin{aligned}
  & \bm{m}_t^{(j)} = (1-\beta_1)\sum_{k=1}^t \beta_1^{t-k} \bm{g}_k^{(j)} \\
  & \bm{v}_t^{(j)} = (1-\beta_2)\sum_{k=1}^t \beta_2^{t-k} (\bm{g}_k^{(j)})^2.
  \end{aligned}
  \end{equation}
  Then,
  \begin{equation}
  \begin{aligned}
  \frac{\vert\bm{m}_t^{(j)}\vert}{\sqrt{\bm{v}_t^{(j)}}+\epsilon} \le & \frac{1-\beta_1}{\sqrt{1-\beta_2}}\cdot \frac{\vert\sum_{k=1}^t \beta_1^{t-k} \bm{g}_k^{(j)}\vert}{\sqrt{\sum_{k=1}^t \beta_2^{t-k} (\bm{g}_k^{(j)})^2}} \\
  \overset{(i)}{\le} & \frac{1-\beta_1}{\sqrt{1-\beta_2}}\cdot \frac{\sum_{k=1}^t \beta_1^{t-k} \vert\bm{g}_k^{(j)}\vert}{\sqrt{\sum_{k=1}^t \beta_2^{t-k} (\bm{g}_k^{(j)})^2}} \\
  \overset{(ii)}{\le}& \frac{1-\beta_1}{\sqrt{1-\beta_2}}\cdot \frac{\sqrt{\sum_{k=1}^t \beta_2^{t-k} (\bm{g}_k^{(j)})^2}\sqrt{\sum_{k=1}^t\frac{\beta_1^{2(t-k)}}{\beta_2^{t-k}}}}{\sqrt{\sum_{k=1}^t \beta_2^{t-k} (\bm{g}_k^{(j)})^2}} \\
  =& \frac{1-\beta_1}{\sqrt{1-\beta_2}}\cdot  \sqrt{\sum_{k=1}^t\frac{\beta_1^{2(t-k)}}{\beta_2^{t-k}}} \\
  \overset{(iii)}{\le}& \frac{1-\beta_1}{\sqrt{1-\beta_2}\sqrt{1-\frac{\beta_1^2}{\beta_2}}},
  \end{aligned}
  \end{equation}
  where $(i)$ holds due to the fact $\vert\bm{a}^{(j)} + \bm{b}^{(j)}\vert \le \vert\bm{a}^{(j)}\vert + \vert\bm{b}^{(j)}\vert $; $(ii)$ holds due to Cauchy-Schiwaz inequality; $(iii)$ holds due to $\beta_1^2 \le \beta_2$.

(2)  The maximal value of $\frac{\vert\bm{m}_t^{(j)}\vert}{\sqrt{\bm{v}_t^{(j)}}}$ is lower bounded by
\begin{equation}
  \frac{1-\beta_1}{\sqrt{1-\beta_2}\sqrt{1-\frac{\beta_1^2}{\beta_2}}} \ge  \frac{1-\beta_1}{{1-\frac{1}{2}(\beta_2 + \frac{\beta_1^2}{\beta_2})}} \ge { \frac{1-\beta_1}{1-\beta_1}} = 1,
\end{equation}
where the first and second inequality reaches the lower bound if and only $\beta_1 = \beta_2$.

When $\beta_1 \le \beta_2$, the maximal value of $\frac{\vert\bm{m}_t^{(j)}\vert}{\sqrt{\bm{v}_t^{(j)}}}$ is upper bounded by
\begin{equation}
  \frac{1-\beta_1}{\sqrt{1-\beta_2}\sqrt{1-\frac{\beta_1^2}{\beta_2}}} \le \frac{1-\beta_1}{\sqrt{1-\beta_2}\sqrt{1-\frac{\beta_1\beta_2}{\beta_2}}} = \frac{\sqrt{1-\beta_1}}{\sqrt{1-\beta_2}}.
\end{equation}

\begin{lemma}
For any random variable $ {\mathcal Z}$ and a constant $C$, there exists
\begin{equation}
  {\mathbb E}[\vert{\rm Sign}({\mathcal Z})- {\rm Sign}(C)\vert] \le \frac{2{\mathbb E}[\vert {\mathcal Z} - C \vert]}{\vert C \vert}.
\end{equation}
\label{lamma_3}
\end{lemma}

\textbf{Proof.} Using Markov'equality, we direct obtain
\begin{equation}
  \begin{aligned}
  {\mathbb E}[\vert{\rm Sign}({\mathcal Z})- {\rm Sign}(C)\vert]  = & 2{\mathbb P}[{\mathbb I}({\rm Sign}({\mathcal Z}) \neq {\rm Sign}(C))] \\
  \le& 2{\mathbb P}[\vert {\mathcal Z} - C \vert \ge \vert C \vert]  \\
  \le& \frac{2{\mathbb E}[\vert {\mathcal Z} - C \vert]}{\vert C \vert}.
  \end{aligned}
\end{equation}

\begin{lemma}
Let $a, b, c >0$ and $0<\alpha<\beta$. If $x\ge ({a+b^{\sfrac{\alpha}{\beta}}})^{\sfrac{1}{\alpha}}$, then $\frac{a}{x^{\alpha}} + \frac{b}{x^{\beta}} \le 1$. The bound is tight up to the factor of 2 since $\frac{({a+b^{\sfrac{\alpha}{\beta}}})^{\sfrac{1}{\alpha}}}{2} \le \max({a}, {b^{\sfrac{1}{\gamma}}}) \le ({a+b^{\sfrac{\alpha}{\beta}}})^{\sfrac{1}{\alpha}} $.
\label{lamma_4}
\end{lemma}

\textbf{Proof.} Let $s=x^\alpha$ and $\gamma=\frac{\beta}{\alpha}\ge 1$, then the inequality becomes
\begin{equation}
\frac{a}{s} + \frac{b}{s^{\gamma}} \le 1
\label{S.18}
\end{equation}

If $s^*$ is a solution of Eq. (\ref{S.18}), it should satisfy
\begin{equation}
s^* \ge \max({a}, {b^{\sfrac{1}{\gamma}}}).
\label{S.19}
\end{equation}

When we set $s_+ = {a+b^{\sfrac{1}{\gamma}}}$, it is easy to verify that
\begin{equation}
\frac{a}{s_+} + \frac{b}{s_+^{\gamma}} = \frac{a}{a+b^{\sfrac{1}{\gamma}}} + \frac{b}{(a+b^{\sfrac{1}{\gamma}})^\gamma} \le \frac{a}{a+b^{\sfrac{1}{\gamma}}}+\frac{b^{\sfrac{1}{\gamma}}}{a+b^{\sfrac{1}{\gamma}}} = 1.
\end{equation}

On the other hand, it is also easy to verify that $s_- = \frac{a+b^{\sfrac{1}{\gamma}}}{2}$ does not satisfy Eq. (\ref{S.19}), which means that $s_+$ is at most a factor of $2$ worse than the smallest solution of Eq. (\ref{S.18}), so $x_+= ({a+b^{\sfrac{\alpha}{\beta}}})^{\sfrac{1}{\alpha}}$ is at most a factor of $2$ worse than the smallest solution of $\frac{a}{x^{\alpha}} + \frac{b}{x^{\beta}} \le 1$.

\subsection{ Proof of Theorem 3.1}

\textbf{Proof.} Following Lemma \ref{lamma_smooth} with $\bm{x}_{t+1}\rightarrow \bm{y}$ and $\bm{x}_{t}\rightarrow \bm{x}$, we have
\begin{equation}
F(\bm{x}_{t+1}) \le  F(\bm{x}_t) + \langle \nabla F(\bm{x}_t), \bm{x}_{t+1} -\bm{x}_t \rangle + \frac{L_0 + L_1\Vert \nabla F(\bm{x}_t) \Vert_2^q}{2}\Vert \bm{x}_{t+1} - \bm{x}_t\Vert_2^2.
\end{equation}
Recalling the update rule $\bm{x}_{t+1} = \bm{x}_t - \gamma \frac{\bm{m}_t}{\sqrt{\bm{v}_t}+\epsilon} = \bm{x}_t - \gamma \frac{\vert \bm{m}_t \vert}{\sqrt{\bm{v}_t}+\epsilon} \circ \frac{\bm{m}_t}{|\bm{m}_t|} = \bm{x}_t - \gamma \bm{u}_t \circ {\rm Sign}(\bm{m}_t) $, we further obtain
\begin{equation}
\begin{aligned}
F(\bm{x}_{t+1}) \le &  F(\bm{x}_t) - \langle \nabla F(\bm{x}_t), \gamma \bm{u}_t \circ {\rm Sign}(\bm{m}_t) \rangle + \frac{(L_0 + L_1\Vert \nabla F(\bm{x}_t) \Vert_2^q)\gamma^2}{2}\Vert \bm{u}_t \Vert_2^2 \\
= & F(\bm{x}_t) - \langle \nabla F(\bm{x}_t), \gamma \bm{u}_t \circ {\rm Sign}(\nabla F(\bm{x}_t)) \rangle + { \left\langle \nabla F(\bm{x}_t), \gamma \bm{u}_t \circ ({\rm Sign}(\nabla F(\bm{x}_t)) - {\rm Sign}(\bm{m}_t) )\right\rangle}  \\
&+\frac{\gamma^2(L_0 + L_1\Vert \nabla F(\bm{x}_t) \Vert_2^q)}{2}\Vert \bm{u}_t \Vert_2^2 \\
\le &  F(\bm{x}_t) - \langle \nabla F(\bm{x}_t), \gamma \bm{u}_t \circ {\rm Sign}(\nabla F(\bm{x}_t)) \rangle + \gamma{R} { \left\langle \nabla F(\bm{x}_t), {\rm Sign}(\nabla F(\bm{x}_t)) - {\rm Sign}(\bm{m}_t) \right\rangle}  \\
&+\frac{\gamma^2 R^2d(L_0 + L_1\Vert \nabla F(\bm{x}_t) \Vert_2^q)}{2},
\end{aligned}
\label{Eq.s_7}
\end{equation}
where the last inequality holds due to $\bm{u}_t^{(j)} \le \frac{1-\beta_1}{\sqrt{1-\beta_2}\sqrt{1-\frac{\beta_1^2}{\beta_2}}}={R}, \forall j\in[d]$ according to {Lemma \ref{lemma.1}}.

Taking the expectation at the $t$ iteration, we obtain
\begin{equation}
\begin{aligned}
{\mathbb E}[{F(\bm{x}_{t+1})}] \le& F(\bm{x}_t) - \gamma\langle \nabla F(\bm{x}_t), {\mathbb E} [\bm{u}_t] \circ {\rm Sign}(\nabla F(\bm{x}_t)) \rangle + \gamma{R} { \left\langle \nabla F(\bm{x}_t), {\mathbb E}[{\rm Sign}(\nabla F(\bm{x}_t)) - {\rm Sign}(\bm{m}_t)] \right\rangle}  \\
&+\frac{\gamma^2 R^2d(L_0 + L_1\Vert \nabla F(\bm{x}_t) \Vert_2^q)}{2} \\
\overset{(i)}{\le} & F(\bm{x}_t) - \gamma\langle \nabla F(\bm{x}_t),  {\mathbb E} [\bm{u}_t] \circ {\rm Sign}(\nabla F(\bm{x}_t)) \rangle  + 2\gamma{R} \; {\mathbb E}[\Vert \bm{m}_t -  \nabla F(\bm{x}_t)\Vert_1] \\
& +\frac{\gamma^2 R^2d(L_0 + L_1\Vert \nabla F(\bm{x}_t) \Vert_2^q)}{2} \\
\overset{(ii)}{\le}& F(\bm{x}_t) - \gamma\langle \nabla F(\bm{x}_t),  {\mathbb E} [\bm{u}_t] \circ {\rm Sign}(\nabla F(\bm{x}_t)) \rangle + 2\gamma R\sqrt{d} \; {\mathbb E}[\Vert \bm{m}_t -  \nabla F(\bm{x}_t)\Vert_2] \\
&+\frac{\gamma^2 R^2d(L_0 + L_1\Vert \nabla F(\bm{x}_t) \Vert_2^q)}{2} \\
\overset{(iii)}{\le}& F(\bm{x}_t) - \gamma\langle \nabla F(\bm{x}_t),  {\mathbb E} [\bm{u}_t] \circ {\rm Sign}(\nabla F(\bm{x}_t)) \rangle + 2\gamma R\sqrt{d} \; {\mathbb E}[\Vert \bm{m}_t -  \nabla F(\bm{x}_t)\Vert_2] \\
&+\frac{\gamma^2 R^2d(L_0 + L_1((1-q)+q{\mathbb E}[\Vert \nabla F(\bm{x}_t) \Vert_2]))}{2},
\end{aligned}
\end{equation}
where $(i)$ holds due to ${\mathbb E}[\vert {\rm Sign}(\nabla F(\bm{x}_t^{(j)})) - {\rm Sign}(\bm{m}_t^{(j)})\vert] \le 2\frac{{\mathbb E}[\vert \nabla F(\bm{x}_t^{(j)}) - \bm{m}_t^{(j)}\vert ]}{\vert\nabla F(\bm{x}_t^{(j)})\vert}(\forall j \in [d])$ according to {Lemma \ref{lamma_3}}; $(ii) $holds owing to the fact $\Vert \bm{a}\Vert_1\le \sqrt{d}\Vert \bm{a}\Vert_2$ for any $\bm{a} \in {\mathbb R}^d$; $(iii)$ holds due to the fact $a^q \le (1-q) + qa$ according to Young's inequality.

Taking expectation from the $1$-st iteration to the $T$-th iteration and then summing them, we have
\begin{equation}
\begin{aligned}
{\mathbb E}[{F(\bm{x}_{t+1})}] \le & F(\bm{x}_1) - \gamma  \sum_{t=0}^{T-1} {\mathbb E} [\Vert \bm{u}_t \circ \nabla F(\bm{x}_t) \Vert_1]  + 2\gamma R\sqrt{d} \; \sum_{t=0}^{T-1}{\mathbb E}[\Vert \bm{m}_t -  \nabla F(\bm{x}_t)\Vert_2] \\
&+\sum_{t=0}^{T-1}\frac{\gamma^2 Rd(L_0 + L_1((1-q)+q\Vert \nabla F(\bm{x}_t) \Vert_2))}{2}.
\end{aligned}
\end{equation}

Rearranging the both sides and applying the facts that ${F(\bm{x}_{t+1})} \ge F^*$, we obtain

\begin{equation}
\begin{aligned}
&\frac{1}{T}\sum_{t=0}^{T-1}{\mathbb E} [\Vert\bm{u}_t \circ \nabla F(\bm{x}_t) \Vert_1]- \frac{\gamma R^2qL_1{d}}{2T}\sum_{t=0}^{T-1} \mathbb E [\Vert\nabla F(\bm{x}_t)\Vert_2] \\
\le & \frac{ F(\bm{x}_{0}) - F^*}{\gamma T}
 + \frac{2R\sqrt{d}}{T}\sum_{t=0}^{T-1} {\mathbb E} [\Vert \bm{m}_t - \nabla F(\bm{x}_t)\Vert_2]
+ \frac{\gamma R^2{d}(L_0+(1-q)L_1)  }{2T}.
\end{aligned}
\label{Eq.s_9}
\end{equation}

Recalling $\bm{m}_t = \beta_1\bm{m}_{t-1} + (1-\beta_1)\bm{g}_t$, we obtain

\begin{equation}
\begin{aligned}
 \bm{m}_t - \nabla F(\bm{x}_t) =&  \left(\beta_1\bm{m}_{t-1} + (1-\beta_1)\bm{g}_t\right) - \nabla F(\bm{x}_t) \\
=& \beta_1(\bm{m}_{t-1} - \nabla F(\bm{x}_{t-1})) +(1-\beta_1)(\bm{g}_t -  \nabla F(\bm{x}_{t}))
 - \beta_1(\nabla F(\bm{x}_t) - \nabla F(\bm{x}_{t-1})).
\end{aligned}
\end{equation}

Utilizing recursion, we further have
\begin{equation}
\begin{aligned}
 \bm{m}_t - \nabla F(\bm{x}_t) =& -\beta_1^t\nabla F(\bm{x}_{0})+ (1-\beta)\sum_{k=1}^t \beta_1^{t-k} (\bm{g}_k -  \nabla F(\bm{x}_k))
  - \sum_{k=1}^t \beta_1^{t-k+1}(\nabla F(\bm{x}_k) - \nabla F(\bm{x}_{k-1})),
\end{aligned}
\end{equation}
where $\bm{m}_1 - \nabla F(\bm{x}_{0}) = -\beta_1 \nabla F(\bm{x}_{0})+(1-\beta_1)(\bm{g}_1-\nabla F(\bm{x}_{0}))$ due to $\bm{m}_0=0$.

Hence,
\begin{equation}
\begin{aligned}
\frac{1}{T}\sum_{t=0}^{T-1} {\mathbb E}\left[\Vert \bm{m}_t - \nabla F(\bm{x}_t) \Vert_2\right]  \le & \underbrace{\frac{1}{T}\sum_{t=0}^{T-1} \beta_1^t\left\Vert \nabla F(\bm{x}_{0})\right\Vert_2}_{{\cal T}_1}+ \underbrace{\frac{1-\beta_1}{T}\sum_{t=0}^{T-1} {\mathbb E}\left[\left\Vert\sum_{k=1}^t \beta_1^{t-k} \left(\bm{g}_k -  \nabla F(\bm{x}_k)\right)\right\Vert_2 \right]}_{{\cal T}_2} \\
 &+  \underbrace{\frac{1}{T}\sum_{t=0}^{T-1} {\mathbb E}\left[\left\Vert \sum_{k=1}^t \beta_1^{t-k+1}(\nabla F(\bm{x}_k) - \nabla F(\bm{x}_{k-1}))\right\Vert_2 \right]}_{{\cal T}_3} \\
\end{aligned}
\label{Eq.s_11}
\end{equation}

In terms of ${\cal T}_1$, we obtain
\begin{equation}
{\cal T}_1 = \frac{1}{T}\sum_{t=0}^{T-1} \beta_1^t\left\Vert \nabla F(\bm{x}_{0})\right\Vert_2 \le \frac{\left\Vert \nabla F(\bm{x}_{0})\right\Vert_2}{T(1-\beta_1)}.
\end{equation}

As for ${\cal T}_2$, we have
\begin{equation}
\begin{aligned}
{\cal T}_2 =&\frac{1-\beta_1}{T}\sum_{t=0}^{T-1} {\mathbb E}\left[\left\Vert\sum_{k=1}^t \beta_1^{t-k} (\bm{g}_k -  \nabla F(\bm{x}_k))\right\Vert_2 \right] \\
\overset{(i)}{\le} & \frac{1-\beta_1}{T}\sum_{t=0}^{T-1}\sqrt{{\mathbb E}\left[\left\Vert\sum_{k=1}^t \beta_1^{t-k} (\bm{g}_k -  \nabla F(\bm{x}_k))\right\Vert_2^2\right]} \\
\overset{(ii)}{=} & \frac{1-\beta_1}{T}\sum_{t=0}^{T-1}\sqrt{\sum_{k=1}^t \beta_1^{2(t-k)} {\mathbb E}\left[\left\Vert\bm{g}_k -  \nabla F(\bm{x}_k)\right\Vert_2^2\right]} \\
\overset{(iii)}{\le} &\frac{1-\beta_1}{T}\sum_{t=0}^{T-1} \sqrt{\sum_{k=1}^t \beta_1^{2(t-k)}(\sigma_0^2 + \sigma_1^2 \Vert \nabla F(x_k) \Vert_2^p)} \\
\overset{(iv)}{\le} &\frac{1-\beta_1}{T}\sum_{t=0}^{T-1} \sqrt{\sum_{k=1}^t \beta_1^{2(t-k)}\sigma_0^2} + \frac{1-\beta_1}{T}\sum_{t=0}^{T-1} \sqrt{\sum_{k=1}^t \beta_1^{2(t-k)}\sigma_1^2 \Vert \nabla F(x_k) \Vert_2^p} \\
\overset{(v)}{\le} &\frac{1-\beta_1}{T}\sum_{t=0}^{T-1} \sqrt{\sum_{k=1}^t \beta_1^{2(t-k)}\sigma_0^2} + \frac{1-\beta_1}{T}\sum_{t=0}^{T-1} \sqrt{\sum_{k=1}^t \beta_1^{2(t-k)}\sigma_1^2\left(\frac{2-p}{2} + \frac{p}{2} \Vert \nabla F(x_k) \Vert_2^2\right)} \\
\overset{(vi)}{\le} &\frac{1-\beta_1}{T}\sum_{t=0}^{T-1} \sqrt{\sum_{k=1}^t \beta_1^{2(t-k)}\sigma_0^2} +\frac{1-\beta_1}{T}\sum_{t=0}^{T-1} \sqrt{\sum_{k=1}^t \frac{(2-p)\sigma_1^2}{2}\beta_1^{2(t-k)}} \\
&+ \frac{1-\beta_1}{T}\sum_{t=0}^{T-1} \sqrt{\sum_{k=1}^t \frac{p}{2} \beta_1^{2(t-k)}\sigma_1^2\Vert \nabla F(x_k) \Vert_2^2} \\
\overset{(vii)}{\le} & \frac{1-\beta_1}{\sqrt{1-\beta_1^2}}\left(\sigma_0 + \sqrt{\frac{2-p}{2}}\sigma_1 \right)
+ \frac{1-\beta_1}{T}\sum_{t=0}^{T-1} \sqrt{\sum_{k=1}^t \frac{p\sigma_1^2}{2} \beta_1^{2(t-k)} \Vert \nabla F(x_k) \Vert_2^2}  \\
\overset{(viii)}{\le} & \frac{1-\beta_1}{\sqrt{1-\beta_1^2}}\left(\sigma_0 + \sqrt{\frac{2-p}{2}} \sigma_1 \right) + (1-\beta_1)\sqrt{\frac{1}{T}\sum_{t=0}^{T-1} \sum_{k=1}^t \frac{p\sigma_1^2}{2} \beta_1^{2(t-k)} \Vert \nabla F(x_k) \Vert_2^2}  \\
\overset{(ix)}{\le} & \frac{1-\beta_1}{\sqrt{1-\beta_1^2}}\left(\sigma_0 + \sqrt{\frac{2-p}{2}}\sigma_1 \right) + (1-\beta_1)\sqrt{ \frac{p\sigma_1^2}{2} \cdot\frac{1}{T}\sum_{t=0}^{T-1} \Vert \nabla F(x_t) \Vert_2^2 \sum_{k=t}^T \beta_1^{2(t-k)}}  \\
{\le} & \frac{1-\beta_1}{\sqrt{1-\beta_1^2}}\left(\sigma_0 + \sqrt{\frac{2-p}{2}}\sigma_1\right) + \frac{\sqrt{p}\sigma_1(1-\beta_1)}{\sqrt{2(1-\beta_1^2)}}\sqrt{ \frac{1}{T}\sum_{t=0}^{T-1} \Vert \nabla F(x_t) \Vert_2^2}  \\
\overset{(x)}{\le} & \frac{1-\beta_1}{\sqrt{1-\beta_1^2}}\left(\sigma_0 + \sqrt{\frac{2-p}{2}}\sigma_1\right)  + \frac{C_0\sqrt{p}\sigma_1(1-\beta_1)}{\sqrt{2(1-\beta_1^2)}} \cdot\frac{1}{T}\sum_{t=0}^{T-1} \Vert \nabla F(x_t) \Vert_2\\
\le & \sqrt{1-\beta_1}\left(\sigma_0 + \sqrt{\frac{2-p}{2}} \sigma_1\right) + {C_0\sigma_1\sqrt{p(1-\beta_1)}} \cdot\frac{1}{T}\sum_{t=0}^{T-1} \Vert \nabla F(x_t) \Vert_2 \\
\end{aligned}
\end{equation}
where  $(i)$ holds due to the fact $({\mathbb E}[Z])^2 \le {\mathbb E}[Z^2] $; $(ii)$ holds owing to ${\mathbb E}[\bm{g}_k -  \nabla F(\bm{x}_k)]=\bm{0} $ according to Assumption A; $(iii)$ holds resulting from ${\mathbb E}\left[\left\Vert\bm{g}_k -  \nabla F(\bm{x}_k)\right\Vert_2^2\right]\le \sigma^2$ according to Assumption C.3; $(iv)$ holds due to the fact $\sqrt{a+b} \le \sqrt{a}+\sqrt{b}$; $(v)$ holds resulting from the fact that $a^p \le \frac{2-p}{2} + \frac{p}{2} a^2, 0\le p \le 2$ according to Young's inequality; $(vi)$ holds due to using the fact $\sqrt{a+b} \le \sqrt{a}+\sqrt{b}$ again;$(vii)$ holds due to the fact  $\sum_i^{T}{a_i}\le \sqrt{T}\sqrt{\sum_i^T a_i^2} $ according to Cauchy-Schwaz inequality; $(viii)$ holds owing to $\sum_{k=1}^t \beta_1^{2(t-k)} \le \frac{1}{1-\beta_1^2}$; $(ix)$ holds thanks to the fact $\sum_{i=1}^n \sum_{j=1}^i f(i,j)g(j)  = \sum_{j=1}^n g(j)\sum_{i=j}^n f(i,j)$; $(x)$ holds because of {Condition 1}.

Now we turn attention to ${\cal T}_3$, \emph{i.e.},
\begin{equation}
\begin{aligned}
{\cal T}_3 = & \frac{1}{T}\sum_{t=0}^{T-1} {\mathbb E}\left[\left\Vert \sum_{k=1}^t \beta_1^{t-k+1}(\nabla F(\bm{x}_k) - \nabla F(\bm{x}_{k-1}))\right\Vert_2 \right] \\
\overset{(i)}{\le} & \frac{1}{T}\sum_{t=0}^{T-1}\sum_{k=1}^t\beta_1^{t-k+1}{\mathbb E}\left[\left\Vert \nabla F(\bm{x}_k) - \nabla F(\bm{x}_{k-1}) \right\Vert_2 \right] \\
\overset{(ii)}{\le} & \frac{1}{T}\sum_{t=0}^{T-1} \sum_{k=1}^t\beta_1^{t-k+1}{\mathbb E}\left[(L_0 + L_1\Vert \nabla F(\bm{x}_{k}) \Vert_2)\Vert\bm{x}_k - \bm{x}_{k-1} \Vert_2 \right] \\
\overset{(iii)}{=}&\frac{1}{T}\sum_{t=0}^{T-1}\sum_{k=1}^t\beta_1^{t-k+1}{\mathbb E}\left[\gamma(L_0 + L_1\Vert \nabla F(\bm{x}_{k}) \Vert_2)\Vert \bm{u}_{t-1} \Vert_2 \right])\\
\overset{(iv)}{\le} & \frac{1}{T}\sum_{t=0}^{T-1} L_1\gamma R\sqrt{d}\sum_{k=1}^t\beta_1^{t-k+1} + \frac{\beta^2L_1\gamma R\sqrt{d}}{T}\sum_{k=1}^t\beta_1^{t-k+1}{\mathbb E}[\Vert \nabla F(\bm{x}_{k}\Vert_2^q)] \\
\le & \frac{\gamma L_0 R \sqrt{d}}{1-\beta} + \frac{\gamma L_1 R\sqrt{d}}{T}\sum_{t=0}^{T-1}\sum_{k=1}^t\beta_1^{t-k+1}((1-q)+q{\mathbb E}[\Vert\nabla F(\bm{x}_{k})\Vert_2]) \\
\overset{(v)}{\le} & \frac{\gamma (L_0+(1-q)L_1) R \sqrt{d}}{1-\beta} + \frac{\gamma qL_1 R\sqrt{d}}{T}\sum_{t=0}^{T-1}\sum_{k=1}^t\beta_1^{t-k+1}{\mathbb E}[\Vert\nabla F(\bm{x}_{k})\Vert_2] \\
\overset{(vi)}{=}& \frac{\gamma (L_0+(1-q)L_1) R \sqrt{d}}{1-\beta_1} + \frac{\gamma qL_1 R\sqrt{d}}{T}\sum_{k=1}^T{\mathbb E}[\Vert\nabla F(\bm{x}_{k})\Vert_2]\sum_{t=k}^T \beta_1^{t-k+1} \\
\le & \frac{\gamma (L_0+(1-q)L_1) R \sqrt{d}}{1-\beta_1} + \frac{\gamma qL_1 R\sqrt{d}}{(1-\beta_1)T}\sum_{t=0}^{T-1}{\mathbb E}[\Vert\nabla F(\bm{x}_{t})\Vert_2] \\
\end{aligned}
\label{Eq.s_14}
\end{equation}
where $(i)$ holds due to the fact $\Vert \bm{a} + \bm{b} \Vert_2 \le \Vert \bm{a} \Vert_2 + \Vert\bm{b}\Vert_2$; $(ii)$ holds owing to Assumption B.3; $(iii)$ holds due to the update rule; $(iv)$ holds depending on $\bm{u}^{(j)} \le \sfrac{1-\beta_1}{\sqrt{1-\beta_2}\sqrt{1-\frac{\beta_1^2}{\beta_2}}} = R$ according to Lemma \ref{lemma.1}; $(v)$ holds thanks to the fact that $a^q \le (1-q) + qa$ according to Young's inequality;  $(vi)$ holds resulting from the fact that $\sum_{i=1}^n\sum_{j=1}^i \bm{a}_{i,j}= \sum_{j=1}^n\sum_{i=j}^n \bm{a}_{i,j}$.

Combining Eq.(\ref{Eq.s_11}) - Eq.(\ref{Eq.s_14}), we have
\begin{equation}
\begin{aligned}
\frac{1}{T}\sum_{t=1}^{T}{\mathbb E}\left[\Vert \bm{m}_t - \nabla F(\bm{x}_t) \Vert_2\right]  \le & \frac{\left\Vert \nabla F(\bm{x}_{0})\right\Vert_2}{T(1-\beta_1)}+ {\sqrt{1-\beta_1}}\left(\sigma_0 + \sqrt{\frac{2-p}{2}} \sigma_1\right) \\
+& {C_0\sigma_1\sqrt{p(1-\beta_1)}} \cdot\frac{1}{T}\sum_{t=0}^{T-1} \Vert \nabla F(x_t) \Vert_2 \\
 +& \frac{\gamma R \sqrt{d} (L_0+(1-q)L_1) }{1-\beta_1} + \frac{\gamma R \sqrt{d} qL_1}{1-\beta_1}\cdot \frac{1}{T}\sum_{t=0}^{T-1}{\mathbb E}[\Vert\nabla F(\bm{x}_{t})\Vert_2]\\
\end{aligned}
\label{Eq.s_15}
\end{equation}

Combining  Eq.(\ref{Eq.s_9}) and Eq.(\ref{Eq.s_15}), we obtain
\begin{equation}
\begin{aligned}
& \frac{1}{T}\left(\sum_{t=0}^{T-1} {\mathbb E} [\Vert\bm{u}_t \circ \nabla F(\bm{x}_t) \Vert_1]- \left(\frac{\gamma R^2dqL_1}{2}+ {{2}C_0R\sqrt{d}\sigma_1\sqrt{p(1-\beta_1)}}+ \frac{2                                                                                                      \gamma R^2{d}qL_1 }{1-\beta_1}\right)\sum_{t=0}^{T-1} \mathbb E [\Vert\nabla F(\bm{x}_t)\Vert_2]\right)  \\
\le & \frac{ F(\bm{x}_{0}) - F^*}{\gamma T} +  \frac{2R\sqrt{d}\left\Vert \nabla F(\bm{x}_{0})\right\Vert_2}{T(1-\beta_1)} + {2R\sqrt{d}\sqrt{1-\beta_1}}\left(\sigma_0+ \sqrt{\frac{2-p}{2}} \sigma_1\right) \\
&+ \frac{2\gamma R^2 d(L_0+(1-q)L_1) }{1-\beta_1} + \frac{\gamma R^2{d}(L_0+(1-q)L_1)  }{2T}.
\end{aligned}
\label{Eq.s_17}
\end{equation}


\subsection{Proof of Corollary 3.2}

\textbf{Proof.} When C.1 holds, it implies $\sigma_1=0$. Also, it is known that $\Vert \nabla F(\bm{x}_t) \Vert_2 \le \Vert \nabla F(\bm{x}_t) \Vert_1$.  Hence,  the conclusion in Theorem 2 can be simplified as

\begin{equation}
\begin{aligned}
 \left(v-\frac{\gamma R^2dqL_1}{2}- \frac{2\gamma R^2{d}qL_1 }{1-\beta_1}\right)\cdot \frac{1}{T}\sum_{t=0}^{T-1} {\mathbb E} [\Vert \nabla F(\bm{x}_t) \Vert_1]
\le & \frac{ F(\bm{x}_{0}) - F^*}{\gamma T} +  \frac{2R\sqrt{d}\left\Vert \nabla F(\bm{x}_{0})\right\Vert_2}{T(1-\beta_1)} \\
& + {2\sqrt{1-\beta_1}R\sqrt{d}}\sigma_0
+ \frac{2\gamma R^2 d \hat{L} }{1-\beta_1} + \frac{\gamma R^2{d}\hat{L}  }{2T},
\end{aligned}
\label{Eq.s_36}
\end{equation}
where $0< v \le \min_{t,j} \bm{u}_t^{(j)}$ and $\hat{L} = L_0+(1-q)L_1$.

Choosing $\gamma=\frac{C_2}{T^{\sfrac{3}{4}}d^{\sfrac{1}{2}}}$ and $1-\beta_1= \frac{C_3}{T^{\sfrac{1}{2}}}$ and  $T \ge (\frac{4C_2 R^2d^{\sfrac{1}{2}}qL_1}{C_3v} +  (\frac{C_2 R^2d^{\sfrac{1}{2}}qL_1}{v})^{\sfrac{1}{3}})^4$, following Lemma \ref{lamma_4}, it holds that
\begin{equation}
\frac{\gamma R^2dqL_1}{2}+ \frac{2\gamma R^2{d}qL_1 }{1-\beta_1} \le \frac{v}{2}.
\end{equation}


Then,  we arrive the conclusion

\begin{equation}
\begin{aligned}
\frac{1}{T}\sum_{t=0}^{T-1} \mathbb E [\Vert\nabla F(\bm{x}_t)\Vert_1] \le& \frac{1}{v}\left(\frac{2(F(\bm{x}_{0})- F(\bm{x}^*))}{C_2T^{\sfrac{1}{4}}} + \frac{4Rd^{\sfrac{1}{2}}\left\Vert \nabla F(\bm{x}_{0})\right\Vert_2}{C_3T^{\sfrac{1}{2}}} \right. \\
+&\left.\frac{4C_3Rd^{\sfrac{1}{2}}\sigma_0}{T^{\sfrac{1}{4}}} +  \frac{4C_2 R^2{d}^{\sfrac{1}{2}}\hat{L}}{C_3T^{\sfrac{1}{4}}} + \frac{C_2R^2d^{\sfrac{1}{2}}\hat{L}}{T^{\sfrac{7}{4}}}\right).
\end{aligned}
\end{equation}

\section{Proof of Corollary 3.3}

\textbf{Proof.} (1)  Choosing $\bar{v} = \min_t \mathbb E [\bm{u}_t^{(j)}] $, we have
\begin{equation}
\begin{aligned}
\sum_{t=0}^{T-1} {\mathbb E} [\Vert\bm{u}_t \circ \nabla F(\bm{x}_t) \Vert_1] =  & \sum_{t=0}^{T-1} \sum_{j=1}^d{\mathbb E} [ \vert\bm{u}_t^{(j)} \nabla F(\bm{x}_t^{(j)}) \vert]  \\
\overset{(i)}{=} & \sum_{t=0}^{T-1} \sum_{j=1}^d{\mathbb E} [ \bm{u}_t^{(j)}] {\mathbb E} [\vert \nabla F(\bm{x}_t^{(j)}) \vert]  \\
\overset{(ii)}{=} & \sum_{t=0}^{T-1}{\mathbb E} [ \bm{u}_t^{(j)}] \sum_{j=1}^d {\mathbb E} [\vert \nabla F(\bm{x}_t^{(j)}) \vert]  \\
= & \sum_{t=0}^{T-1} {\mathbb E} [ \bm{u}_t^{(j)}] {\mathbb E} [ \Vert (\nabla F(\bm{x}_t)) \Vert_1] \\
\overset{(iii)}{\ge} & \bar{v}\sum_{t=0}^{T-1} {\mathbb E} [ \Vert (\nabla F(\bm{x}_t)) \Vert_1]  \\
\overset{(iv)}{=}& \frac{\bar{v}\sqrt{d}}{C_1}\sum_{t=0}^{T-1}{\mathbb E} [\Vert (\nabla F(\bm{x}_t)) \Vert_2],
\end{aligned}
\end{equation}
where $(i)$ holds due to $\bm{u}_t^{(j)}$ and $\vert \nabla F(\bm{x}_t^{j}) \vert$ are mutually independent; $(ii)$ holds owing to applying Condition 2; $(iii)$ holds due to the condition $\bar{v} \le \min_{t} {\mathbb E} [ \bm{u}_t^{(j)}]$ ; $(iv)$ holds depending on Condition 3.

Then, we simplify the conclusion in Theorem \ref{lemma_convergence} as
\begin{equation}
\small
\begin{aligned}
 & \left(\bar{v}-\left(\frac{\gamma C_1R^2\sqrt{d}qL_1}{2} + {{2}C_0C_1R\sigma_1\sqrt{p(1-\beta_1)}} + \frac{2\gamma C_1 R^2\sqrt{d}qL_1 }{1-\beta_1}\right) \right)\cdot \frac{1}{T}\sum_{t=0}^{T-1} {\mathbb E} [\Vert \nabla F(\bm{x}_t) \Vert_2] \\
\le & \frac{ C_1(F(\bm{x}_{0}) - F^*)}{\gamma T\sqrt{d}} +  \frac{2C_1R\left\Vert \nabla F(\bm{x}_{0})\right\Vert_2}{T(1-\beta_1)}
 + {2C_1\sqrt{1-\beta_1}R}\hat{\sigma}
+ \frac{2\gamma C_1R^2 \sqrt{d} \hat{L} }{1-\beta_1} + \frac{\gamma C_1 R^2\sqrt{d}\hat{L}}{2T}.
\end{aligned}
\label{Eq.s_41}
\end{equation}

Choosing  $\gamma=\frac{C_2}{T^{\sfrac{3}{4}}d^{\sfrac{1}{2}}}$, $1-\beta_1= \frac{C_3}{T^{\sfrac{1}{2}}}$ and $T \ge (\frac{4C_2 R^2qL_1}{C_3\bar{v}} +  {4}C_0C_1\sqrt{C_3}R\sigma_1\sqrt{p} +  (\frac{C_2 R^2qL_1}{\bar{v}})^{\sfrac{1}{3}})^4$, following Lemma \ref{lamma_4}, it holds that
\begin{equation}
\frac{\gamma C_1R^2\sqrt{d}qL_1}{2} + {{2}C_0C_1R\sigma_1\sqrt{p(1-\beta_1)}} + \frac{2\gamma C_1 R^2\sqrt{d}qL_1 }{1-\beta_1} \le \frac{\bar{v}}{2}.
\end{equation}


Then, we reformulate Eq. (\ref{Eq.s_41}) as

\begin{equation}
\small
\hspace{-10pt}
\begin{aligned}
\frac{1}{T}\sum_{t=0}^{T-1} \mathbb E [\Vert\nabla F(\bm{x}_t)\Vert_2] \le& \frac{C_1}{\bar{v}}\left(\frac{2( F(\bm{x}_{0})- F(\bm{x}^*))}{C_2T^{\sfrac{1}{4}}d^{\sfrac{1}{2}}} + \frac{4R\left\Vert \nabla F(\bm{x}_{0})\right\Vert_2}{C_3T^{\sfrac{1}{2}}}
+ \frac{4C_3R\hat{\sigma}}{T^{\sfrac{1}{4}}} +  \frac{4C_2 R^2\hat{L}}{C_3T^{\sfrac{1}{4}}} + \frac{C_2R^2\hat{L}}{T^{\sfrac{7}{4}}} \right).
\end{aligned}
\end{equation}

(2) Using generalized Young's inequality, we minimize the bottle-neck terms to obtain the lowerest bound on the right hand of Eq. (\ref{Eq.s_41}), \emph{i.e.},
\begin{equation}
\begin{aligned}
& \frac{ C_1(F(\bm{x}_{0}) - F^*)}{\gamma T \sqrt{d}} + {2C_1\sqrt{1-\beta_1}R}\hat{\sigma}
+ \frac{2C_1\gamma R^2 \sqrt{d} \hat{L} }{1-\beta_1}  \\
\ge& \left(\frac{4 C_1(F(\bm{x}_{0}) - F^*)}{\gamma T \sqrt{d}}\right)^{\sfrac{1}{4}}\cdot \left({4C_1\sqrt{1-\beta_1}R}\hat{\sigma}\right)^{\sfrac{1}{2}} \cdot \left( \frac{8C_1\gamma R^2 \sqrt{d} \hat{L} }{1-\beta_1}\right)^{\sfrac{1}{4}} \\
= &  \frac{512^{\sfrac{1}{4}} C_1R\hat{\sigma}^{\sfrac{1}{2}}\hat{L}^{\sfrac{1}{4}}(F(\bm{x}_{0}) - F^*)}{T^{\sfrac{1}{4}}},
\end{aligned}
\end{equation}
where the lowest bound achieved if and only if $\frac{4 C_1(F(\bm{x}_{0}) - F^*)}{\gamma T \sqrt{d}} = {4C_1\sqrt{1-\beta_1}R}\hat{\sigma}$  and $ {4C_1\sqrt{1-\beta_1}R}\hat{\sigma} = \frac{8C_1\gamma R^2 \sqrt{d} \hat{L} }{1-\beta_1}$, and we further obtain
\begin{equation}
\begin{aligned}
&\gamma = \frac{(F(\bm{x}_{0}) - F^*)^{\sfrac{3}{4}}}{2^{\sfrac{1}{4}}T^{\sfrac{3}{4}}d^{\sfrac{1}{2}}R\hat{\sigma}^{\sfrac{1}{2}}\hat{L}^{\sfrac{1}{4}}}, \\
\\
& \beta_1 = 1 - \frac{2^{\sfrac{1}{2}}\hat{L}^{\sfrac{1}{2}}(F(\bm{x}_{0}) - F^*)^{\sfrac{1}{2}}}{T^{\sfrac{1}{2}}\hat{\sigma}}.
\end{aligned}
\end{equation}

When it is chosen $T \ge (\frac{4C_0C_1\hat{C}_3^{\sfrac{1}{2}}p^{\sfrac{1}{2}}R\sigma_1}{v}+\frac{4C_1\hat{C_2} R^2qL_1}{\hat{C}_3v} +  (\frac{C_1\hat{C}_2 R^2qL_1}{v})^{\sfrac{1}{3}})^4$ where $\hat{C}_2=\frac{(F(\bm{x}_{0}) - F^*)^{\sfrac{3}{4}}}{2^{\sfrac{1}{4}}R\hat{\sigma}^{\sfrac{1}{2}}\hat{L}^{\sfrac{1}{4}}}$ and $C_3=\frac{2^{\sfrac{1}{2}}\hat{L}^{\sfrac{1}{2}}(F(\bm{x}_{0}) - F^*)^{\sfrac{1}{2}}}{\hat{\sigma}}$, following Lemma \ref{lamma_4}, it holds that
\begin{equation}
\frac{\gamma C_1R^2\sqrt{d}qL_1}{2} + {{2}C_0C_1R\sigma_1\sqrt{p(1-\beta_1)}} + \frac{2\gamma C_1 R^2\sqrt{d}qL_1 }{1-\beta_1} \le \frac{v}{2}.
\end{equation}

Then, we reformulate Eq. (\ref{Eq.s_41}) as

\begin{equation}
\begin{aligned}
\frac{1}{T}\sum_{t=0}^{T-1} \mathbb E [\Vert\nabla F(\bm{x}_t)\Vert_2] \le& \frac{C_1} {\bar{v}} \left( \frac{512^{\sfrac{1}{4}} R\hat{\sigma}^{\sfrac{1}{2}}\hat{L}^{\sfrac{1}{4}}(F(\bm{x}_{0}) - F^*)}{T^{\sfrac{1}{4}}} + \frac{2C_1\hat{C}_3R\left\Vert \nabla F(\bm{x}_{0})\right\Vert_2}{T^{\sfrac{3}{4}}} +  \frac{  \hat{C_2}  R^2\hat{L}  }{T^{\sfrac{7}{4}}}\right).
\end{aligned}
\end{equation}

\end{document}